\documentclass[lettersize,journal]{IEEEtran}
\usepackage{amsmath,amsfonts}
\usepackage{algorithmic}
\usepackage{algorithm}
\usepackage{array}
\usepackage[caption=false,font=normalsize,labelfont=sf,textfont=sf]{subfig}
\usepackage{textcomp}
\usepackage{stfloats}
\usepackage{url}
\usepackage{verbatim}
\usepackage{graphicx}
\usepackage{cite}
\usepackage{threeparttable}

\usepackage{booktabs}
\usepackage{nicematrix}
\usepackage{multirow}
\usepackage{color}
\usepackage{pifont}
\usepackage{amssymb}
\usepackage{bm}
\usepackage{tablefootnote}
\usepackage{hyperref}

\usepackage{bbm} 
\definecolor{revisecolor}{RGB}{00, 00, 00}
\newcommand\luowei[1]{\textcolor{revisecolor}{#1}}
\newcommand\ouyangfan[1]{\textcolor{revisecolor}{#1}}

\definecolor{revisecolor2}{RGB}{00, 00, 00}
\newcommand\revise[1]{\textcolor{revisecolor2}{#1}}

\begin{document}

\title{ProtoDCS: Towards Robust and Efficient Open-Set Test-Time Adaptation for Vision-Language Models}

\author{Wei~Luo\textsuperscript{*,1,2},
        Yangfan~Ou\textsuperscript{*,3},
        Jin~Deng\textsuperscript{1},
        Zeshuai~Deng\textsuperscript{3},
        Xiquan~Yan\textsuperscript{4},
        Zhiquan~Wen\textsuperscript{\dag,3},
        and~Mingkui~Tan\textsuperscript{\dag,2}
\thanks{Wei Luo and Jin Deng are with the \textsuperscript{1}South China Agricultural University. \textit{Email:{\{cswluo, jindsmu\}@gmail.com}}. 
Mingkui Tan is with the \textsuperscript{2}Pazhou Laboratory. \textit{Email:{mingkuitan@scut.edu.cn}}. 
Yangfan Ou, Zeshuai Deng, and Zhiquan Wen are with the \textsuperscript{3}South China University of Technology. \textit{Email:{swouyangfan@163.com}, {sedengzeshuai}@mail.scut.edu.cn}, sewenzhiquan@gmail.com. 
Xiquan Yan is with the \textsuperscript{4}Hunan Normal University. \textit{Email:{yanxiquan@126.com}}. 
Wei Luo is also with the \textsuperscript{2}Pazhou Laboratory.}%
\thanks{This work was supported in the part by
National Natural Science Foundation of China (Grant No.82550116), Joint Funds of the National Natural Science Foundation of China (Grant No.U24A20327), the Young Scholar Project of Pazhou Lab (Grant No.PZL2021KF0021).}%
\thanks{* Authors contributed equally. \dag Corresponding authors.}%
\thanks{This paper was produced by the IEEE Publication Technology Group. They are in Piscataway, NJ.}%
\thanks{Manuscript received April 19, 2021; revised August 16, 2021.}}

\markboth{Journal of \LaTeX\ Class Files,~Vol.~14, No.~8, August~2021}%
{Shell \MakeLowercase{\textit{et al.}}: A Sample Article Using IEEEtran.cls for IEEE Journals}



\maketitle

\begin{abstract}
Large-scale Vision-Language Models (VLMs) exhibit strong zero-shot recognition, yet their real-world deployment is challenged by distribution shifts. While Test-Time Adaptation (TTA) can mitigate this, existing VLM-based TTA methods operate under a closed-set assumption, failing in open-set scenarios where test streams contain both covariate-shifted in-distribution (csID) and out-of-distribution (csOOD) data. This leads to a critical difficulty: the model must discriminate unknown csOOD samples to avoid interference while simultaneously adapting to known csID classes for accuracy.
Current open-set TTA (OSTTA) methods rely on hard thresholds for separation and entropy minimization for adaptation. These strategies are brittle, often misclassifying ambiguous csOOD samples and inducing overconfident predictions, and their parameter-update mechanism is computationally prohibitive for VLMs.  
To address these limitations, we propose Prototype-based Double-Check Separation (ProtoDCS), a robust framework for OSTTA that effectively separates csID and csOOD samples, enabling safe and efficient adaptation of VLMs to csID data. Our main contributions are: (1) a novel double-check separation mechanism employing probabilistic Gaussian Mixture Model (GMM) verification to \revise{mitigate the brittleness of rigid thresholding}; and (2) an evidence-driven adaptation strategy utilizing uncertainty-aware loss and efficient prototype-level updates, mitigating overconfidence and reducing computational overhead.
Extensive experiments on CIFAR-10/100-C and Tiny-ImageNet-C demonstrate that ProtoDCS achieves state-of-the-art performance, significantly boosting both known-class accuracy and OOD detection metrics. 
\textit{Code will be available at \url{https://github.com/O-YangF/ProtoDCS}.}
\end{abstract}


\section{Introduction}

\IEEEPARstart{P}{retrained} large-scale vision-language models (VLMs), such as CLIP~\cite{radford2021CLIP}, ALIGN~\cite{jia2021ALIGN}, and the BLIP series~\cite{li2022blip, li2023blip2}, have demonstrated remarkable zero-shot capabilities across various downstream tasks~\cite{yu2022coca}. However, their deployment in dynamic real-world environments faces significant challenges due to distribution shifts~\cite{hendrycks2021many,recht2019imagenet}. To address this problem, test-time adaptation (TTA)~\cite{wang2021tent,eata@22icml,zanella2025realistic, niu2023SAR} has emerged as a promising solution, enabling models to adapt using only unlabeled target data. Therefore, recent studies have tended to combine VLMs and TTA to achieve a more powerful adaptation effect~\cite{shu2022TPT,karmanov2024efficient,ctpt@24iclr,wang2025tapt,dai2025free}, addressing tasks ranging from VQA \cite{liu2024question} to video recognition \cite{zeng2025exploring}.

\begin{figure}[t]
    \centering
    \includegraphics[width=\columnwidth]{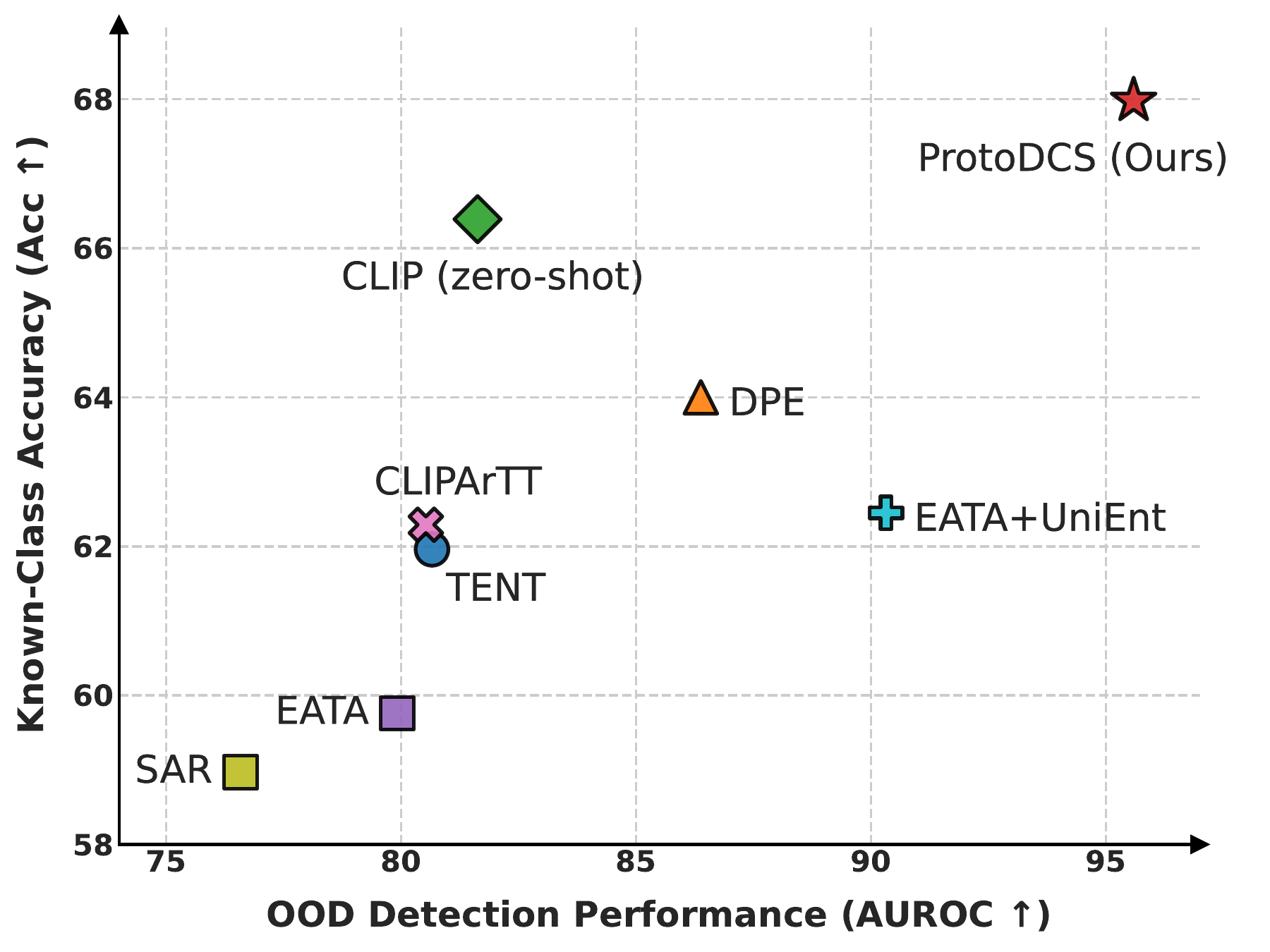}
    \caption{\textbf{Performance comparison on CIFAR-10-C.} Our ProtoDCS simultaneously achieves the highest known-class accuracy (ACC) and the best OOD detection performance (AUROC), significantly outperforming all baseline methods. The top-right region indicates better performance.}
    \label{fig:compare}
    \vspace{-5mm}
\end{figure}

Yet, existing VLM-based TTA methods are mainly designed for the \textit{covariate‑shifted in‑distribution} (csID) scenario, in which every test class is known beforehand~\cite{zhang2024DPE,zhang2024boostadapter,shu2022TPT,karmanov2024efficient,ctpt@24iclr,wang2025tapt,dai2025free}. This closed‑set assumption, however, breaks down in safety‑critical applications such as autonomous driving, where the data stream inevitably contains \textit{covariate-shifted out-of-distribution} (csOOD) instances drawn from previously unseen classes~\cite{guan2024UniEnt, gong2023wisdom, yu2024stamp, wang2025effortless}. For instance, an autonomous vehicle might encounter unusual construction equipment (csOOD) while driving in foggy conditions (covariate shift). Compulsorily classifying such previous unseen objects into seen categories could lead to severe traffic risks. This open-set condition~\cite{hendrycks2017baseline, liang2018odin, liu2020energy} presents a fundamental difficulty for TTA: the model must simultaneously perform two conflicting tasks on the unlabeled test stream -- \textit{discriminate csOOD data to avoid interference, while adapting to csID data to maintain accuracy}.

Recent open-set TTA (OSTTA) methods are primarily designed for small-scale networks~\cite{guan2024UniEnt,gong2023wisdom,zhou2025motta}. 
Among them, UniEnt~\cite{guan2024UniEnt} separates csID and csOOD samples by their similarity to non-adapted known class prototypes of the source domain, 
while~\cite{gong2023wisdom} leverages the absolute difference of prediction confidence between the original and updated model for separation and MoTTA~\cite{zhou2025motta} filters samples based on representation stability under pruning. 
Despite these advancements, these methods suffer from several key limitations when applying them to VLM-based TTA. 
\textbf{1) Brittle Separation from Hard Thresholding.} These methods universally depend on rigid thresholds (e.g., fixed similarity or confidence difference) for csID/csOOD separation. This paradigm is intrinsically mismatched with the highly compact and well-structured embedding space of VLMs~\cite{understandingCRL@20icml}, where hard thresholds are prone to misclassifying ambiguous boundary samples. Consequently, erroneous gradients from misclassified csOOD data corrupt the representations of known classes, distort the decision boundary, and ultimately impair csID accuracy.
\textbf{2) Unsafe Adaptation from Entropy Minimization.} The adaptation relies heavily on entropy minimization (EM), which blindly sharpens predictions without discerning their correctness. This not only fosters overconfidence on noisy or boundary samples but, more critically, disrupts the inherent calibration of pre-trained VLMs~\cite{calibrateVLM@24icml,calibrateVLA@24eccv}. The corruption of this crucial property destabilizes the adaptation process and leads to performance degradation.
Moreover, these methods intrinsically rely on gradient-based parameter updates, making their computational cost prohibitive for large-scale VLMs. These fundamental limitations collectively hinder their direct and effective extension to the VLM regime.

Given these limitations, developing an effective VLM-based OSTTA model must confront two major challenges: 
1) \textbf{Towards robust csID/csOOD separation beyond hard thresholding}.
The model requires a separation mechanism that moves beyond brittle, hard thresholds. It must adaptively handle the ambiguous region where csID and csOOD distributions overlap, preventing misclassified csOOD samples from corrupting the adaptation process and thus preserving csID accuracy.
2) \textbf{Towards safe and efficient adaptation for VLMs}.
The adaptation strategy must satisfy a dual requirement. First, it should replace entropy minimization with an objective that explicitly models predictive uncertainty to prevent overconfident predictions on noisy or boundary samples. Second, to be feasible for large-scale VLMs, the strategy must be highly efficient, ideally avoiding computationally expensive gradient computations through the entire backbone.

To address these challenges, we propose \textbf{P}rototype-based \textbf{D}ouble-\textbf{C}heck \textbf{S}eparation (\textbf{ProtoDCS}), a novel framework that systematically resolves the OSTTA difficulties through a cohesive design centered on reliable separation and efficient adaptation. 
\textbf{To achieve robust separation beyond hard thresholds (Challenge 1)}, ProtoDCS employs a progressive two-stage verification mechanism. The \textit{First-Check} stage computes an openness score and applies dual-threshold filtering to segregate samples: confident samples ($S_{open}(x) < \Theta_a$) populate a diversity-aware visual cache for robust prototype construction, while trustworthy samples ($S_{open}(x) < \Theta_b$) proceed for further assessment. The \textit{Final-Verification} stage then leverages a Gaussian Mixture Model (GMM) on (re-evaluated) openness scores to provide probabilistic, adaptive separation that naturally handles distribution shifts and ambiguous boundary cases. 
\textbf{To enable safe and efficient adaptation for VLMs (Challenge 2)}, ProtoDCS operates exclusively at the prototype level, keeping the VLM backbone frozen. For trustworthy samples, we perform {evidence-driven temporary prototype optimization} using residual updates, where an uncertainty-aware loss explicitly models both aleatoric and epistemic uncertainty to prevent overconfidence, replacing overconfidence-prone entropy minimization. Only samples confirmed by both verification stages contribute to {asymmetric prototype evolution}: textual prototypes are updated gradually via cumulative moving average to maintain semantic stability, while visual prototypes undergo more selective updates through the diversity-aware cache mechanism. 
This closed-loop system ensures that prototype evolution is driven exclusively by high-fidelity csID samples, enabling effective domain adaptation while preserving discriminability against csOOD instances. We show the performance comparison of ProtoDCS with other methods  in Fig.~\ref{fig:compare}.

In summary, motivated by identifying three fundamental limitations in current OSTTA research when applied to VLMs -- the hard threshold trap, the overconfidence problem, and the computational inefficiency, 
our work addresses these issues to make the following concrete contributions: 
\begin{itemize}
    \item We propose ProtoDCS, a probabilistic double-check separation mechanism with GMM verification, which \revise{mitigates the brittleness of rigid thresholding} in VLMs' compact embedding space, thereby enabling adaptive and reliable csID/csOOD separation under distribution shifts.
    \item We develop an evidence-driven adaptation strategy with a novel uncertainty-aware loss, directly countering the overconfidence issue of entropy minimization, which ensures safe model updates through better-calibrated predictions in open-set scenarios.
    \item We design a lightweight prototype-level update framework that keeps the VLM backbone frozen, effectively overcoming the computational bottleneck of gradient backpropagation, and achieving highly efficient adaptation without performance loss.


\end{itemize}

\section{Related Work}
\label{sec:related_work}
    
\subsection{Close-set Test-Time Adaptation}
Test-Time Adaptation (TTA) aims to enhance model robustness against distribution shifts by using only unlabeled test data. Early methods primarily focused on closed-set scenarios, developing strategies around entropy minimization~\cite{wang2021tent}, consistency regularization~\cite{memo@22nips}, pseudo-labeling~\cite{zhang2023non-parametric}, \luowei{and collaborative model-data co-adaptation~\cite{modelanddataTTA26tcsvt}}. While effective within their intended setting, these methods operate under the critical and often unrealistic assumption that all test samples belong to known classes, making them inherently vulnerable in open-set environments where unknown classes appear during testing. 
The advent of large Vision-Language Models (VLMs) introduced additional complexity to TTA. While some approaches have explored VLM adaptation through prompt tuning~\cite{shu2022TPT},\luowei{~\cite{dualprompttuningTTA26tcsvt}}, lightweight adapters~\cite{gao2024clipadapter}, \luowei{or Bayesian modeling~\cite{bayesianTTA25CVPR}}, these methods typically require computationally expensive gradient backpropagation through the entire backbone. 
Recently, more efficient alternatives have emerged, including stem-layer adaptation~\cite{shin2024ltta}, embedding-space adjustments \cite{zanella2024mta}, \luowei{feature refinement~\cite{meanshiftTTA26tcsvt}}, and training-free prototype adjustment~\cite{iwasawa2021T3A}, yet they remain constrained by the closed-set assumption. This fundamental limitation renders them ineffective for real-world deployment where unknown categories inevitably appear alongside distribution shifts. 
We address these dual challenges by performing lightweight adaptation exclusively at the prototype level, eliminating the need for backbone gradient computations while simultaneously maintaining robustness against open-set interference. Our approach preserves the efficiency benefits of representation-level adaptation while extending its applicability to the more realistic and challenging open-set scenario.


\subsection{Open-Set Test-Time Adaptation}
\ouyangfan{Unlike offline open-set domain adaptation (OSDA) which assumes access to labeled source data~\cite{tong2025reserve},} Open-Set TTA (OSTTA) presents a more realistic but challenging scenario where test streams contain both covariate-shifted in-distribution (csID) and out-of-distribution (csOOD) samples. The primary risk in this setting is ``negative adaptation'', where conventional TTA methods misclassify csOOD samples with high confidence, progressively corrupting the model's representations~\cite{guan2024UniEnt, chen2022CoTTA}. Existing OSTTA methods generally adopt two strategies: sample filtering based on confidence metrics~\cite{gong2023wisdom},~\luowei{\cite{cliptta25nips}} or model stability~\cite{zhou2025motta}, and unified optimization objectives designed for joint csID adaptation and csOOD rejection~\cite{guan2024UniEnt, dai2025free}. 
However, these approaches face significant limitations when applied to VLMs. Most critically, they rely on brittle hard thresholds for csID/csOOD separation, which proves particularly problematic in the highly compact and well-structured embedding space of VLMs~\cite{understandingCRL@20icml}. In such spaces, where class representations are naturally clustered with narrow inter-class margins, fixed thresholds inevitably misclassify ambiguous boundary samples. Furthermore, existing methods fail to fully leverage the rich multimodal information inherent in VLMs, and their gradient-based update mechanisms remain computationally burdensome for large-scale models~\luowei{\cite{openirt25nips}}. 
We introduce a comprehensive solution through a novel double-check separation mechanism. We replace hard thresholds with a probabilistic Gaussian Mixture Model that adapts to the evolving test distribution, enabling robust separation even in VLMs' compact embedding spaces. By operating entirely at the prototype level and incorporating both visual and textual modalities, our framework achieves computationally efficient adaptation while maintaining the cross-modal alignment crucial for VLM performance. 




\subsection{Uncertainty Calibration in TTA}

Mitigating overconfidence during test-time adaptation is crucial for model reliability. A predominant strategy in early TTA methods is entropy minimization (EM)~\cite{wang2021tent,niu2023SAR,eata@22icml}. While effective in boosting prediction confidence, EM blindly sharpens the output distribution without discerning correctness, inevitably leading to overconfident yet inaccurate predictions on noisy or ambiguous samples. 
To alleviate this issue, subsequent research has branched broadly into three directions. 
One line of work seeks to restrict the application of EM to a more reliable subset of test samples. For instance, EATA-C~\cite{eatac@2025pami} leverages prediction consistency between full and pruned networks as a reliability criterion, while DeYo~\cite{deyo@24iclr} selects samples exhibiting significant prediction discrepancy against input destruction. 
Alternatively, another direction incorporates regularization terms to calibrate the EM process. POEM~\cite{poem@24nips} aligns the test-time entropy with a source-domain prior to prevent drastic distribution shifts. In the context of Vision-Language Models (VLMs), methods like C-TPT~\cite{ctpt@24iclr} and O-TPT~\cite{sharifdeen2025o-tpt} introduce a dispersion loss to encourage textual prototypes to maintain maximal separation, thereby preserving the model's inherent discriminative capability. 
Beyond EM, a more fundamental approach is to explicitly model predictive uncertainty within the loss function. TEA~\cite{tea@24cvpr} retains model uncertainty by minimizing an energy score that represents multi-class compatibility, whereas COME~\cite{tang2024come} and EDL~\cite{edl@18nips} adopt subjective logic to explicitly construct and optimize belief and uncertainty masses. 
Despite these advancements, a critical gap persists in VLM-based OSTTA. Current sample-selection heuristics struggle in VLMs' compact representation space, while prevailing uncertainty methods—though addressing epistemic uncertainty—largely ignore the inherent aleatoric uncertainty in test data. Departing from entropy minimization, we introduce evidence-driven optimization that explicitly models both uncertainty types in a unified loss function. Grounded in evidential deep learning, our approach naturally penalizes overconfidence on ambiguous csID or unfamiliar csOOD samples, ensuring safe adaptation that preserves VLMs' intrinsic well-calibrated nature—crucial for robust open-set performance.


\section{Preliminary}
\label{sec:preliminary}

\noindent\textbf{Contrastive Language-Image Pre-Training (CLIP).}
CLIP~\cite{radford2021CLIP} is a vision-language model composed of a visual encoder $E_v(\cdot)$ and a text encoder $E_t(\cdot)$, which map images and textual prompts into a shared feature space $\mathbb{R}^d$. For an input image $x$, its embedding is computed as $\bm{f}_v = E_v(x)$, while each class prompt $T_k$ is encoded as $\bm{f}_{t,k} = E_t(T_k)$. Classification is then performed by computing the cosine similarity between $\bm{f}_v$ and each $\bm{f}_{t,k}$ followed by a softmax over the similarities: 
\begin{equation}
\label{eq:clip_softmax}
P(y = c_k \mid x) = \frac{\exp\left( \text{cos}(\bm{f}_v, \bm{f}_{t,k}) / \tau \right)}{\sum_{j=1}^{K} \exp\left( \text{cos}(\bm{f}_v, \bm{f}_{t,j}) / \tau \right)},
\end{equation}
where \(\text{cos}(\bm{a}, \bm{b}) = \frac{\bm{a} \cdot \bm{b}}{\|\bm{a}\| \|\bm{b}\|}\) denotes cosine similarity, and \(\tau\) is the temperature parameter.  
After pretraining, CLIP maps images and text into a shared embedding space, enabling zero-shot prediction across diverse tasks, such as zero-shot classification and cross-modal retrieval.

\noindent\textbf{Open-Set Test-Time Adaptation (OSTTA).}
Let \( \mathcal{X}_s = \{x_i\}_{i=1}^{N_s} \) denote the input samples from a labeled source domain, and let \( \mathcal{Y}_s = \{1, \dots, K\} \) denote the corresponding label space consisting of \(K\) known classes. 
A model \( f_{\theta_0} \), pre-trained on \( (\mathcal{X}_s, \mathcal{Y}_s) \), is deployed in a test-time setting where only unlabeled target inputs \( \mathcal{X}_t = \{x_j\}_{j=1}^{N_t} \) are accessible. Unlike traditional closed-set settings, the target domain may contain both in-distribution samples from known classes under domain shift (referred to as \textbf{csID}) and out-of-distribution samples from unseen classes, also under domain shift (referred to as \textbf{csOOD}). 
The goal of OSTTA is to adapt the model to the target stream \( \mathcal{X}_t \) while maintaining performance on known classes and avoiding degradation caused by updates driven by csOOD samples.


\begin{figure*}[t]
    \centering
    \includegraphics[width=\textwidth]{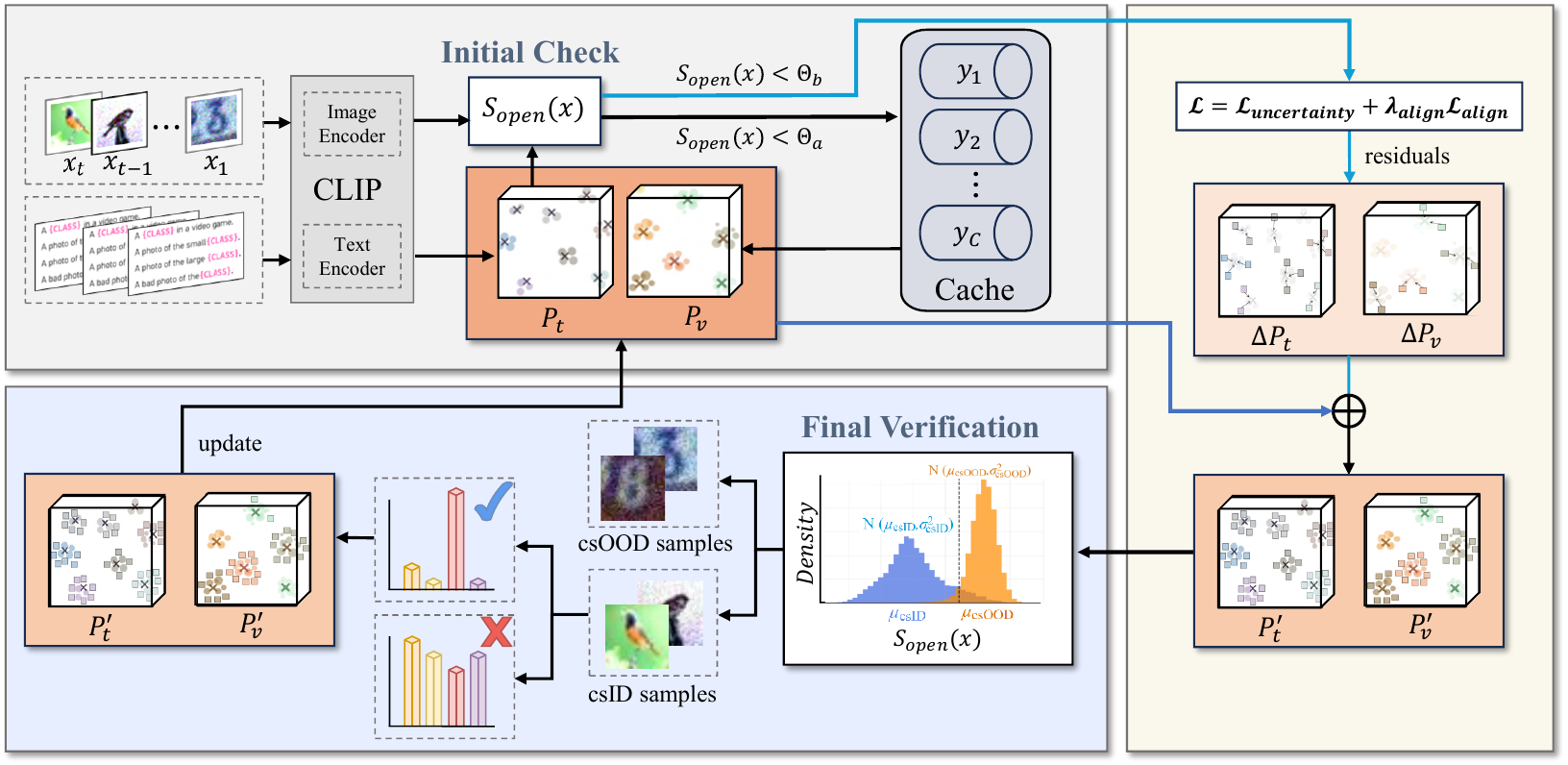}
    \caption{\textbf{Overview of the proposed ProtoDCS.} 
    For an incoming unlabeled sample $x$, ProtoDCS first computes its initial openness score \(S_{open}(x)\). The \textbf{First-Check} stage uses this score for an initial separation: \textit{confident samples} (\(S_{open}(x) < \Theta_a\)) populate the diversity-aware visual cache to build visual prototypes \(\bm{P}_v\), while \textit{trustworthy samples} (\(S_{open}(x) < \Theta_b\)) undergo evidence-driven optimization to generate temporary prototypes \(\bm{P}'_v, \bm{P}'_t\) for a refined openness assessment. The \textbf{Final-Verification} stage then leverages a GMM on the (re-evaluated) openness scores for all samples to make a probabilistic csID/csOOD decision. Only the sample confirmed as csID through this double-check process contribute to the final prototype evolution, forming a robust closed-loop adaptation system free from csOOD corruption.
    }
    \label{fig:ProtoDCS_workflow}
    \vspace{-5mm}
\end{figure*}

\section{Prototype-based Double-Check Separation}
\label{sec:method}

\noindent\textbf{Problem Statement.}
Existing VLM-based TTA methods~\cite{shu2022TPT, zhang2024DPE} typically assume a closed-set setting and rely on unsupervised objectives such as entropy minimization. However, when unknown-class samples appear during adaptation, two critical issues arise: 
(1) Erroneous updates from misidentified csOOD samples. Unknown‑class samples can be easily mistaken as reliable data, leading to gradient updates that corrupt the model’s representations and degrade performance. 
(2) Overconfidence from entropy minimization. Entropy‑based TTA forces the model to assign excessively confident posteriors, which is especially harmful for noisy or boundary samples, further undermining adaptation robustness~\cite{tang2024come}. 
These two issues reflect the core conflict in OSTTA: the need to simultaneously reject csOOD samples and safely adapt to csID data. To resolve this conflict, a dedicated framework must provide (1) a robust mechanism for filtering out csOOD samples before update, and (2) an uncertainty‑aware adaptation objective that prevents overconfidence and enables safe prototype evolution.


\noindent\textbf{Method Overview.} 
In this section, we present ProtoDCS (Prototype‑based Double‑Check Separation), a framework designed to overcome the above issues through a cohesive, prototype‑centric design. ProtoDCS addresses three key limitations of existing OSTTA methods when applied to VLMs:
(1) Brittle separation due to hard thresholds, ill‑suited to VLMs’ compact embedding space;
(2) Unsafe adaptation caused by overconfidence‑prone entropy minimization;
and (3) Computational inefficiency from gradient updates through the entire backbone. 
As illustrated in Fig.~\ref{fig:ProtoDCS_workflow}, ProtoDCS introduces two core innovations:
(1) A progressive two‑stage verification mechanism that \revise{augments rigid thresholds with an adaptive, probabilistic separation} using a Gaussian Mixture Model (GMM), thereby handling ambiguous boundary samples robustly.
(2) An evidence‑driven adaptation strategy that employs an uncertainty‑aware loss to explicitly model both aleatoric and epistemic uncertainty, preventing overconfidence while enabling lightweight prototype‑level updates without back‑propagation through the frozen backbone.
By ensuring that only high‑fidelity csID samples contribute to prototype evolution, ProtoDCS forms a closed‑loop system that effectively reconciles reliable csOOD filtering with safe and efficient adaptation. 
The following subsections detail each component; the complete algorithm is summarised in supplementary materials (SMs).

\subsection{First-Check: Visual Cache Construction and Trustworthy Sample Selection}
\label{sec:initial_assessment}
A primary challenge in open-set TTA is that indiscriminate adaptation to all test samples can corrupt the model's representations when csOOD data is present. To address this, the First-Check stage is designed to achieve two critical objectives: (1) to construct robust visual prototypes exclusively from high-fidelity in-distribution (csID) samples, ensuring a pure and discriminative feature space; and (2) to identify a broader set of trustworthy samples that, while potentially ambiguous, warrant further inspection for adaptation, thereby balancing robustness with adaptation recall. 
To realize these objectives, this stage operates in two sequential steps: it first quantifies each sample's affinity to known classes via an openness score, then employs a dual-threshold filtering strategy to segregate samples for their respective roles.

\subsubsection{\textbf{Openness Score: A Measure of Familiarity} }

The openness score serves as a metric to estimate the likelihood of a sample $x$ belonging to a known class. The intuition is that a sample with high similarity to any known class prototype is more likely to be in-distribution. Formally, we extract the visual embedding $\bm{f}_v = E_v(x)$ and compare it with the textual prototypes $\bm{P}_t = [\bm{f}_{t,1}, \cdots, \bm{f}_{t,K}]\in\mathbb{R}^{d\times K}$. The openness score is computed as:
\begin{equation}
\label{eq:os_init}
S_{open}(x) = 1 - \max_{k \in \{1,\dots,K\}} z_k,
\end{equation}
with $z_k$ the output logit of CLIP: 
\begin{equation}
\label{eq:z}
z_k =  \frac{\bm{f}_v \cdot \bm{f}_{t,k}}{\|\bm{f}_v\| \|\bm{f}_{t,k}\|},
\end{equation}
where $\lVert \cdot \rVert$ is the $l_2$ norm. A score approaching 0 indicates high similarity to a known class, while a score near 1 suggests a higher probability of being csOOD. Additionally, the initial predicted label $c_k = \arg\max(P(y|x))$ is obtained to guide the subsequent cache update process.

\subsubsection{\textbf{Dual-Threshold Filtering: Balancing Safety and Recall} }
Relying on a single threshold for separation is often brittle, as it fails to account for the ambiguous region where csID and csOOD distributions overlap. We therefore introduce a dual-threshold strategy to manage this trade-off, directing samples to different processing paths based on their confidence level.

\paragraph{\textbf{Cache Admission ($\Theta_a$): A Strict Gate for Prototype Purity}}
\label{sec:diversity_aware}
To build a reliable foundation for visual prototypes, we maintain a class-wise visual cache $\mathcal{C}_v = \{\mathcal{Q}_k\}_{k=1}^K$, where each queue $\mathcal{Q}_k$ (of length $l$) stores representative features for class $k$. A sample $x$ with predicted class $k$ is admitted as a confident sample $x_{conf}$ into $\mathcal{Q}_k$ only if $S_{open}(x) < \Theta_a$, where $\Theta_a$ is a class-agnostic dynamic threshold. To ensure the cache remains both diverse and high-quality, we enforce the following update policy (see Algorithm 2 in SMs for detailed operations):
\begin{itemize}
\item If $x$ exhibits high cosine similarity ($>\tau_{sim}$) to an existing sample in $\mathcal{Q}_k$ but possesses higher quality (i.e., lower aleatoric uncertainty), it replaces the existing sample.
\item If $x$ is dissimilar to all samples in $\mathcal{Q}_k$, it is added if the queue is not full; otherwise, it replaces the sample with the highest aleatoric uncertainty.
\end{itemize}
The visual prototype for class $k$ is subsequently computed as the mean of its queue: $\bm{f}_{v,k} = \frac{1}{|\mathcal{Q}_k|} \sum_{\bm{f} \in \mathcal{Q}_k} \bm{f}.$

\paragraph{\textbf{Trustworthy Sample Selection ($\Theta_b$): A Looser Gate for Potential Adaptation}} 
A stringent $\Theta_a$ ensures prototype purity but may discard valuable, slightly noisier csID samples. To alleviate this loss of informative data, we employ a looser threshold $\Theta_b$ (where $\Theta_b > \Theta_a$). Samples satisfying $S_{open}<\Theta_b$ are classified as trustworthy samples $x_{trust}$. While not pristine enough for direct cache admission, these samples are passed to the next stage for a more refined assessment, allowing the model to explore a wider range of potential adaptations without compromising the core prototypes.

In practice, $\Theta_a$ and $\Theta_b$ are set dynamically as the $30th$ (Q(0.3)) and $60th$ (Q(0.6)) percentiles, respectively, of the openness scores within a sliding window $\mathcal{W}_c$. This adaptive mechanism ensures the filtering strategy remains responsive to the evolving test stream.

\begin{figure}[t!]
    \vspace{-4mm}
    
    \centering
    \subfloat[\small{Entropy Minimization}]{
        \includegraphics[width=0.43\columnwidth]{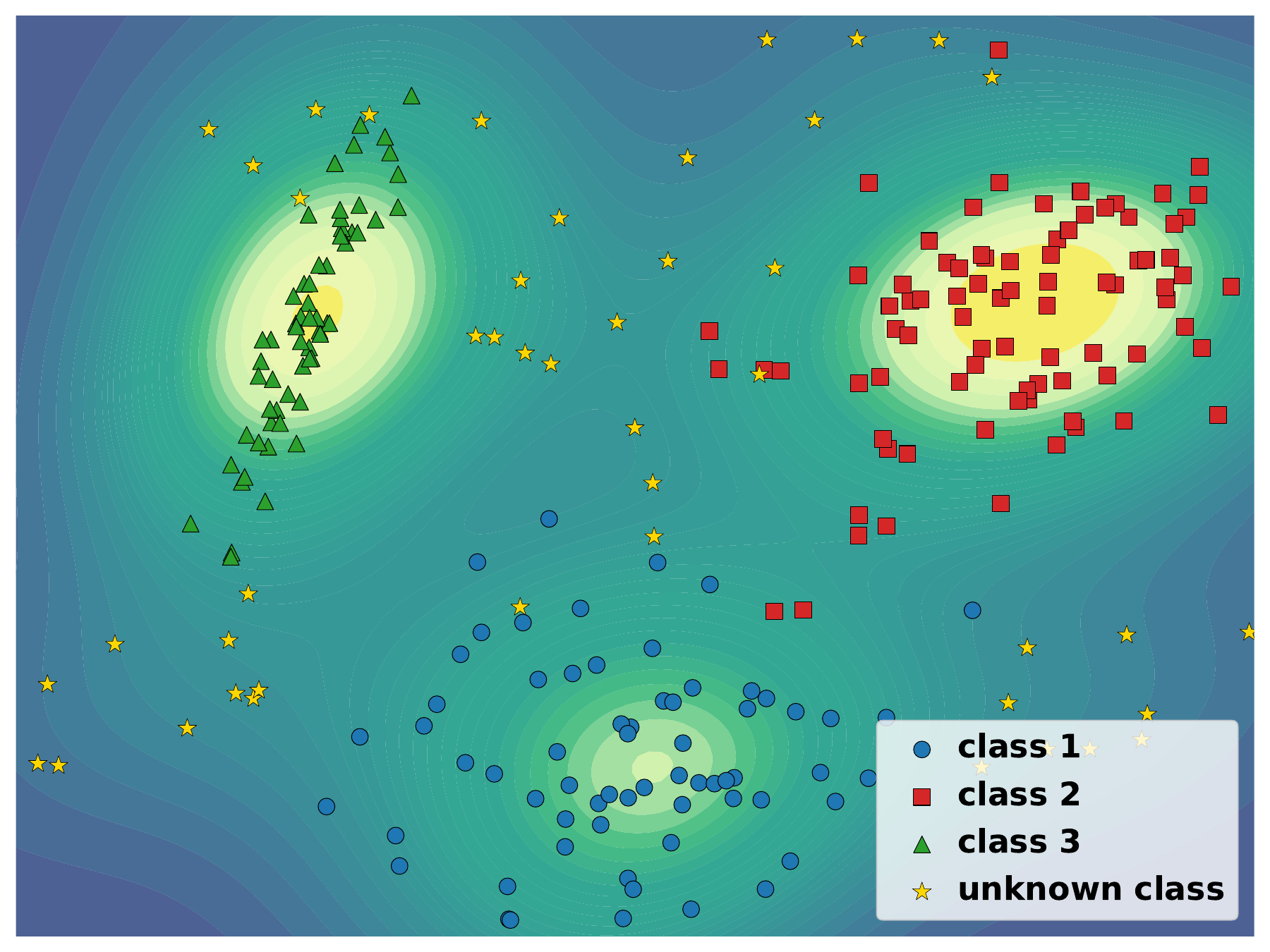}
        \label{fig:entropy_vs_evident-l}
    }
    \subfloat[\small{Evidence-Driven(Ours)}]{
        \includegraphics[width=0.51\columnwidth]{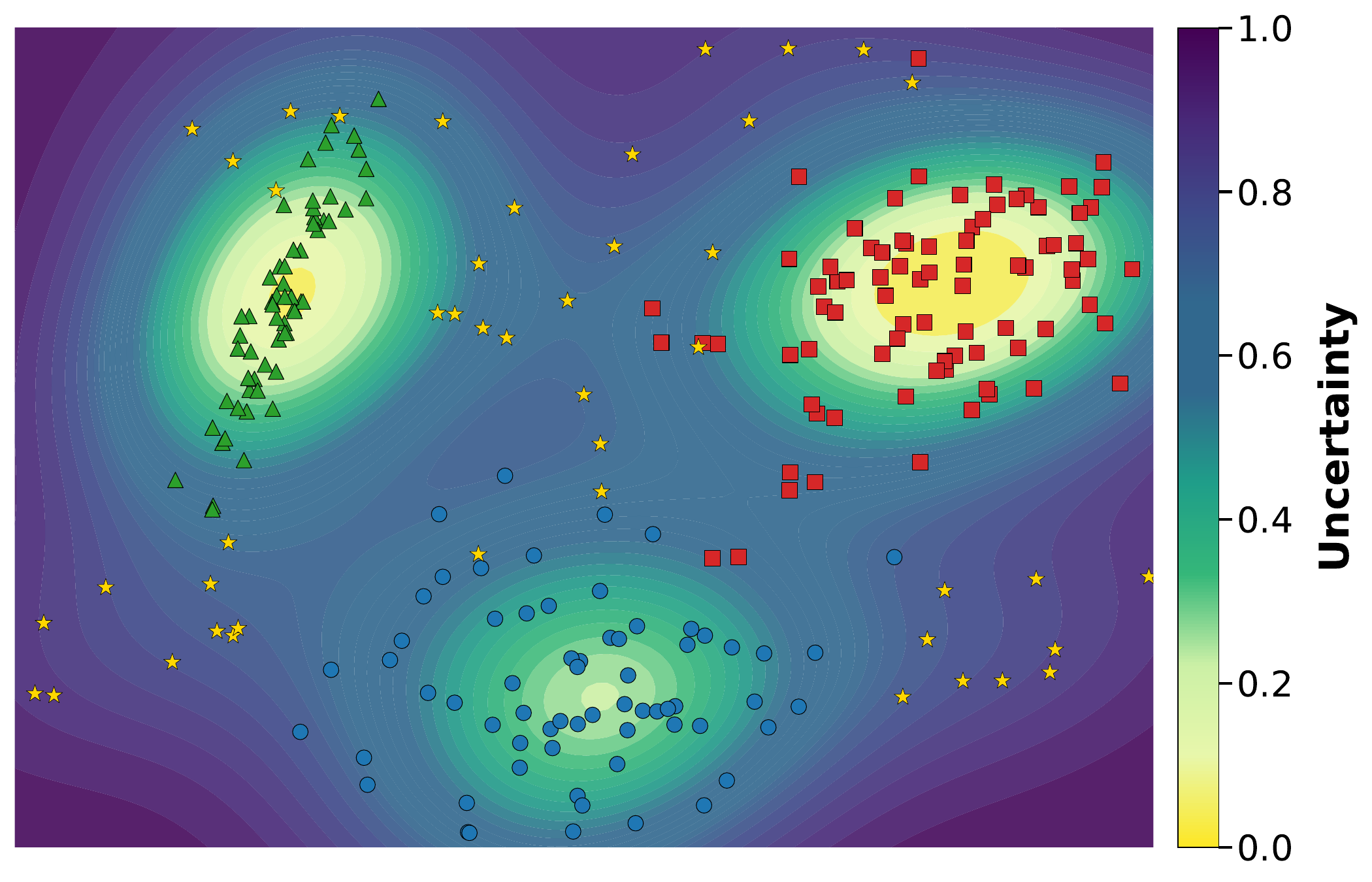}
        \label{fig:entropy_vs_evident-r}
    }
    \caption{
        \textbf{Uncertainty Landscapes in Feature Space.} 
        We compare the uncertainty surface constructed by (a) Standard Entropy Minimization and (b) our Evidence-Driven Uncertainty-Aware (EDUA) loss. 
        The background color intensity denotes the uncertainty level derived from the loss function, where \textbf{Yellow/Light} indicates lower uncertainty and \textbf{Purple/Dark} indicates higher uncertainty.
        Crucially, Entropy Minimization creates a deceptive landscape where csOOD samples fall into low-uncertainty regions. In contrast, ProtoDCS correctly maps the csOOD and boundary regions to high uncertainty (dark background), effectively filtering out risky samples during prototype evolution.
    }
    \vspace{-5mm}
    \label{fig:entropy_vs_evident}
\end{figure}

\subsection{Evidence-driven Temporary Prototype Generation}
\label{sec:diff_optimization}
The \textit{trustworthy samples} identified in the First-Check stage constitute a valuable yet potentially risky resource for adaptation. While they likely contain beneficial information for domain alignment, their ambiguous nature means direct prototype updating could progressively corrupt the model's representations. Therefore, this module aims to \textbf{simulate} adaptation effects on these samples through a safe, evidence-driven optimization process that generates temporary prototypes, thereby avoiding irreversible changes to the global model state until a sample's validity is confirmed

\subsubsection{\textbf{Residual-based Prototype Update: A Safe Simulation}} 
To achieve non-destructive adaptation, we update prototypes not by direct modification but by optimizing learnable residual matrices. Specifically, for a trustworthy sample $x_{trust}$, we introduce residual matrices $\Delta \bm{P}_t$ and $\Delta \bm{P}_v$, initialized to zero and associated with the frozen textual prototypes $\bm{P}_t$ and visual prototypes $\bm{P}_v$, respectively. The temporary prototypes are then computed as:
\begin{equation}
\label{eq:tempo_protos}
    \bm{P}'_t= \text{Norm}(\bm{P}_t + \Delta \bm{P}_t),~~
    \bm{P}'_v= \text{Norm}(\bm{P}_v + \Delta \bm{P}_v),
\end{equation}
where $\text{Norm}(\cdot)$ denotes $\ell_2$ normalization. This residual formulation allows the prototypes to adaptively adjust to the sample while keeping the backbone network frozen. These temporary prototypes serve to re-assess the openness score of $x_{trust}$ and used to output the class prediction if $x_{trust}$ passes the second-round verification and is determined as a csID sample in the subsequent stage.
It is worth noting that $\Delta \bm{P}_t$ is computed for all csID textual prototypes, whereas $\Delta \bm{P}_v$ is only updated for csID classes with non-empty queues in the visual cache $\mathcal{C}_v$, ensuring computational efficiency and focus on established classes.

\subsubsection{\textbf{Uncertainty-Aware Optimization: Preventing Overconfidence}} The residual matrices are optimized by minimizing a composite loss function that jointly enforces prediction confidence and cross-modal consistency:
\begin{equation}
\label{eq:total_loss_condensed_revised}
\min \mathcal{L} = \mathcal{L}_{\text{uncertainty}} + \lambda_{\text{align}} \mathcal{L}_{\text{align}},
\end{equation}

\paragraph{\textbf{Uncertainty Loss ($\mathcal{L}_{\text{uncertainty}}$)}} 
Standard entropy minimization (EM) often yields overconfident predictions even for ambiguous or OOD samples~\cite{tang2024come, sharifdeen2025o-tpt}. To better calibrate uncertainty, we instead adopt evidential learning~\cite{edl@18nips,kendall2017uncertainties}, which models the adapted logits $\bm{z} \in \mathbb{R}^K$ as evidence for a Dirichlet distribution. Specifically, we compute the evidence vector as $\bm{\alpha} = \mathrm{ReLU}(\bm{z}) + 1$, and define the evidence-driven uncertainty-aware (EDUA) loss to explicitly minimize both:
\begin{itemize}
    \item \textbf{Aleatoric Uncertainty (AU):} Captures data ambiguity by measuring the expected prediction entropy.
    \item \textbf{Epistemic Uncertainty (EU):} Reflects the model’s lack of knowledge via the total evidence magnitude.
\end{itemize}
The total uncertainty loss is defined as:
\begin{align}
\label{eq:uncertainty_loss}
\mathcal{L}_{\text{uncertainty}} &= \lambda_{AU}\text{AU}(x) + \text{EU}(x),
\end{align}
where $\lambda_{AU}$ is the balance coefficient and:
\begin{align}
\label{eq:au}
\text{AU}(x) = - \sum_{k=1}^K \frac{\alpha_k}{S} \left( \psi(\alpha_k + 1) - \psi(S + 1) \right), \text{EU}(x) &= \frac{K}{S},
\end{align}
with $S=\sum_{k=1}^K \alpha_k$ and $\psi$ is the digamma function. 

Minimizing this uncertainty-aware loss drives the model to express confidence only when supported by strong class evidence, while preserving uncertainty in ambiguous cases, thereby achieving better-calibrated uncertainty estimates on challenging samples. Crucially, this uncertainty loss is applied exclusively to \textit{trustworthy samples}, ensuring robust prototype refinement without influence by csOOD-like instances. 
\luowei{In contrast to the EM loss, which relies on relative differences among prediction logits, the EDUA loss quantifies sample uncertainty based on absolute logit values.  
This property facilitates effective identification of both genuine csOOD and hard csID samples. 
In addition, an in-depth discriminability analysis of EDUA is deferred to SMs.}

\paragraph{\textbf{Alignment Loss ($\mathcal{L}_{\text{align}}$)}} 
To prevent semantic drift between modalities during adaptation, we introduce an alignment loss based on the InfoNCE objective. This term encourages that the updated visual prototypes $\bm{P}'_v$ and textual prototypes $\bm{P}'_t$ to remain semantically consistent in the shared embedding space:
\begin{equation}
    \mathcal{L}_{\text{align}} = \text{InfoNCE}(\bm{P}'_v, \bm{P}'_t).
\end{equation}
This auxiliary objective acts as a regularizer, mitigating the risk of prototype divergence caused by noisy or incomplete supervision.

\subsubsection{\textbf{Integration and Subsequent Flow}}
After a single gradient descent step minimization of the total loss $\mathcal{L}$, the optimized residuals $\Delta \bm{P}_t$ and $\Delta \bm{P}_v$ are used to compute the temporary prototypes via Eq.~(\ref{eq:tempo_protos}). These temporary prototypes are subsequently employed to re-evaluate the openness score of $x_{trust}$, providing a refined input for the GMM-based Final-Verification in Section~\ref{sec:final_classification}. Crucially, the temporary prototypes are discarded unless $x_{trust}$ is definitively confirmed as a csID sample in the final verification, ensuring that prototype evolution remains robust and free from csOOD contamination.

\subsection{Final-Verification: Probabilistic Separation with GMM}
\label{sec:final_classification}

While the First-Check stage provides an initial sample filtering, its dependence on threshold-based separation proves inadequate for handling distribution shifts and ambiguous boundary cases. To overcome the brittleness of hard thresholds, the Final-Verification stage serves as the definitive decision stage, \revise{augmenting the initial rigid thresholding with a probabilistic framework} that adapts to the evolving characteristics of the test stream. We employ a two-component Gaussian Mixture Model (GMM) to characterize the openness score distribution, enabling robust, context-aware separation between csID and csOOD samples by naturally accommodating uncertainty in the decision process.

The GMM framework models the openness scores as arising from two underlying distributions corresponding to csID and csOOD samples. Formally, represented as:
\begin{equation}
\label{eq:gmm_model_revised}
\mathcal{G}(S_{open}) = \sum_{j=1}^{2} \phi_j \mathcal{N}(S_{open} | \mu_j, \sigma_j^2),
\end{equation}
where $\phi_j$, $\mu_j$, and $\sigma_j^2$ represent the mixture weight, mean, and variance of the $j$-th component, respectively. For component identification, we assign the component with higher mean openness score to represent csOOD samples, reflecting their characteristic of lower similarity to known classes.

In implementation, we maintain a sliding window $\mathcal{W}_g$ of recent openness scores to capture the local distribution dynamics of the target stream. For each incoming sample $x$, we compute its final openness score using the temporary prototypes (if classified as trustworthy in First-Check) or retain the initial score (otherwise). The GMM parameters are initialized using K-means clustering on the sliding window data, ensuring stable convergence during the expectation-maximization process. This adaptive initialization strategy allows the model to automatically adjust to the current test distribution without requiring manual parameter tuning. See SMs for more details.

The definitive classification is determined by the posterior probability:
\begin{equation}
\label{eq:gmm_ood_classification_revised}
P(c_{csOOD} | S_{\text{open}}(x)) > \Theta_p,
\end{equation}
where $\Theta_p = 0.5$ serves as the default confidence threshold. To ensure estimation reliability, we apply GMM-based verification only when sufficient evidence is available, specifically when at least 100 openness scores have been accumulated in $\mathcal{W}_g$. This safeguard prevents unreliable decisions during the initial adaptation phase; 
\luowei{for extremely skewed test streams where the GMM fitting fails, a robust fallback threshold $\Theta_c$ (dynamically updated as the GMM component midpoint) is employed -- see SMs for details.}


\subsection{Model Prediction and Prototype Evolution}
The Final-Verification stage provides definitive csID/csOOD separation, establishing a foundation for reliable model adaptation. This module completes the ProtoDCS pipeline by leveraging confirmed csID samples for robust prediction and controlled prototype evolution, thereby enabling incremental domain adaptation while preserving model integrity against distribution shifts and semantic drift.

\subsubsection{\textbf{Dual-Source Prediction}} 
\label{sec:decision} 
Traditional VLM-based prediction typically relies solely on textual similarity, which may not fully capture the visual characteristics of the target domain. To enhance classification robustness, we introduce a dual-source prediction strategy that synergistically combines information from both textual and visual modalities, leveraging their complementary strengths for improved decision-making in open-set scenarios. 
The prediction framework integrates similarity measures from both prototype spaces:
\begin{equation}
\label{eq:l_final_computation}
c_k = \arg \max (P(y|x) + \mathcal{A}(\bm{f}_v, \bm{P}'_v)),
\end{equation}
where $P(y|x)$ represents the standard textual similarity computed using the adapted textual prototypes $\bm{P}'_t$, while the affinity-based correction term $\mathcal{A}(\cdot)$ incorporates visual guidance:
\begin{equation}
\label{eq:affinity}
\mathcal{A}(\bm{f}_v, \bm{P}'_v)_k = \alpha \cdot \exp(-\beta (1 - \text{cos}(\bm{f}_v, \bm{p}'_{v,k}))),
\end{equation}
Here, $\alpha$ and $\beta$ are scaling hyperparameters that balance the contribution of the visual affinity term. This formulation ensures that predictions benefit from both the semantic grounding of language and the distribution-aware characteristics of visual features, particularly valuable when dealing with domain-shifted samples that may exhibit visual patterns not fully captured by textual descriptions alone~\cite{zhang2024DPE}.

\subsubsection{\textbf{Asymmetric Prototype Update} }
\label{sec:selective_evolution} 
Effective test-time adaptation requires careful balance between responsiveness to distribution shifts and stability against noisy updates. To achieve this balance, we employ an asymmetric update strategy that respects the distinct characteristics of textual and visual modalities, enabling prompt adaptation to visual domain shifts while maintaining semantic stability.

For textual prototypes, we adopt a cumulative moving average (CMA) scheme that facilitates gradual, stable adaptation:
\begin{equation}
\label{eq:cma_update_revised}
\bm{P}_t \leftarrow (1 - \frac{1}{n}) \bm{P}_{t} + \frac{1}{n} \bm{P}'_t,
\end{equation}
where $n$ represents the total number of csID-confirmed samples observed. To further enhance robustness, this update is gated by a dynamic quality threshold, ensuring that only the most reliable csID samples contribute to the evolution of the global prototypes (see SMs for details). 
This conservative approach acknowledges the relative stability of textual semantics derived from class names, ensuring that semantic knowledge accumulated during pretraining is preserved while allowing controlled adaptation to domain-specific differences.

In contrast, visual prototypes undergo more selective updates to accommodate the potentially rapid shifts in visual distributions. 
A sample $x$ contributes to visual prototype evolution only if it satisfies multiple criteria: 
(1) passes the double-check separation ($S_{open}(x) < \Theta_b$ in First-Check and $P(c_{csOOD} | S_{\text{open}}(x)) < \Theta_p$ in Final-Verification), 
and (2) meets the diversity and quality requirements for cache admission as specified in Section~\ref{sec:initial_assessment}. 
This stringent filtering ensures that visual prototypes evolve based exclusively on high-fidelity samples, preventing corruption from boundary csID and ambiguous csOOD instances while enabling responsive adaptation to genuine distribution shifts in the visual domain.

The asymmetric design rationale stems from the fundamental differences in how textual and visual representations respond to domain shifts. While visual features often exhibit significant distribution changes across domains, textual semantics grounded in class names remain comparatively stable. By tailoring update strategies to these modality-specific characteristics, ProtoDCS achieves effective domain adaptation without compromising the cross-modal alignment essential for VLM performance.

\section{Experiment}
\label{sec:experiment}

In this section, we comprehensively evaluate the performance of ProtoDCS across varying architectures and datasets so as to answer the following key concerns about ProtoDCS:
\begin{itemize}
    \item Does the two-stage verification mechanism bring robust csID/csOOD separation and safe adaptation?
    \item Do the components of ProtoDCS achieve the claimed effects?
    \item Is ProtoDCS robust to hyperparameter selection?
    \item How does the computation efficiency of ProtoDCS perform considering its involvement of the two-stage verification mechanism?
    \item How does ProtoDCS perform in close-set TTA settings?
\end{itemize}

\subsection{Experimental Setup}
\label{sec:exp}

\noindent\textbf{{Datasets and Scenarios}}.
Following previous studies~\cite{guan2024UniEnt,gong2023wisdom}, we evaluate ProtoDCS on three standard OSTTA benchmarks -- CIFAR-10-C, CIFAR-100-C, and Tiny-ImageNet-C~\cite{tinyimagenet}. Each of them contains 15 types of corruption with 5 severity levels. 
For CIFAR-10-C and CIFAR-100-C, we use their 10,000 test images as csID data and 26,032 images from SVHN-C as csOOD data. For Tiny-ImageNet-C, we use the 10,000 validation images as csID data and 2,000 images from ImageNet-O-C as csOOD data. 
SVHN-C and ImageNet-O-C respectively contain the same 15 types of corrupted images from the original SVHN~\cite{svhn} and ImageNet-O~\cite{imagenet-o} datasets.
In our experiments, all corruptions are applied at the highest severity level (level 5), creating a challenging test stream with imbalanced csID/csOOD ratios to highlight our method's robustness in unstable environments.

\noindent\textbf{Evaluation Protocol}.
We adopt a rigorous online \luowei{non-episodic} adaptation protocol where the model adapts to a continuous data stream, sample by sample, without any parameter reset. This setup highlights our method's suitability for low-latency, real-world applications. Performance is assessed using four key open-set metrics: \textbf{Acc} for known-class accuracy, \textbf{AUROC} for the model's ability to distinguish csID from csOOD samples, \textbf{FPR@TPR95} for a stricter assessment of OOD detection, and \textbf{OSCR} \cite{dhamija2018reducing}, a holistic metric evaluating overall open-set classification performance.

\noindent\textbf{{Baselines}}.
We compare ProtoDCS against three groups of methods. \textbf{1) Standard TTA Methods:} \textit{TENT}~\cite{wang2021tent}, \textit{SAR}~\cite{niu2023SAR}, and \textit{EATA}~\cite{eata@22icml}.
\textbf{2) VLM-based TTA Methods:} \textit{TPT}~\cite{shu2022TPT}, \textit{C-TPT}~\cite{ctpt@24iclr}, \textit{DiffTPT}~\cite{difftpt}, 
\textit{CLIPArTT}~\cite{clipartt}, \luowei{\textit{BCA}~\cite{bayesianTTA25CVPR}}
\textit{TDA}~\cite{karmanov2024efficient}, and \textit{DPE}~\cite{zhang2024DPE}.
\textbf{3) Open-Set TTA Methods:}
\textit{UniEnt}~\cite{guan2024UniEnt}, \textit{STAMP}~\cite{yu2024stamp}, \luowei{and \textit{CLIPTTA}~\cite{cliptta25nips}}. 
We provide detailed descriptions of these methods in the Supplementary Materials.

\noindent\textbf{{Implementation Details}}.
We use the pre-trained CLIP model with ViT-B/16 as the main backbone, and ResNet-50 for supplementary evaluation. The backbone remains frozen. For adaptation, we use the AdamW optimizer to update textual and visual residuals for a single step. On CIFAR-100-C, the learning rates are set to $2.5 \times 10^{-4}$ for text and $7.5 \times 10^{-3}$ for visual residuals. The visual cache size is set to 5 per class. For separation, the thresholds $\Theta_a$ and $\Theta_b$ are dynamically set to the 30th and 60th percentiles of a 100-sample sliding window. The GMM verification uses a window of $100$ samples with a probability threshold of $0.5$. For prediction, the scaling factors are $\alpha=0.5$ and $\beta=9.5$. $\lambda_{align}$ and $\lambda_{AU}$ are respectively set to 0.2 and 1 for prototype residual updates. We report the average results over 15 corruption types \luowei{on three random runs}.

\begin{table*}[htbp]
    \centering
    \setlength{\tabcolsep}{6pt}
    \caption{Results of different methods on CIFAR benchmarks using the CLIP ViT-B/16 backbone. $\uparrow$ indicates that larger values are better, and vice versa. All values are percentages (\%). 
    Bold indicates the best.}
    \label{tab:cifar_benchmarks_c10_c100}
    \begin{threeparttable}
    \begin{tabular}{@{}lcccccccc@{}}
            \toprule
            \multicolumn{1}{c}{\multirow{2}{*}[-0.4ex]{Method}}
            & \multicolumn{4}{c}{CIFAR-10-C} & \multicolumn{4}{c}{CIFAR-100-C} \\
            
            \cmidrule(lr){2-5} \cmidrule(lr){6-9} 

            & Acc$\uparrow$ & AUROC$\uparrow$ & FPR@TPR95$\downarrow$ & OSCR$\uparrow$
            & Acc$\uparrow$ & AUROC$\uparrow$ & FPR@TPR95$\downarrow$ & OSCR$\uparrow$ \\
            
            \midrule
            CLIP (ViT)& 66.39 & 81.63 & 63.82 & 13.89 & 37.14 & 55.42 & 91.40 & 18.87 \\
            
            \midrule
            TENT& 61.96 & 80.66 & 67.07 & 53.70 & 33.03 & 67.45 & 86.26 & 25.08 \\
            SAR& 58.97 & 76.59 & 73.91 & 49.81 & 25.15 & 54.69 & 89.12 & 14.11 \\
            EATA& 59.76 & 79.92 & 67.76 & 52.41 & 32.16 & 66.55 & 86.04 & 24.21 \\

            \midrule
            TPT& 64.50 & 79.06 & 72.55 & 55.62 & 34.39 & 42.18 & 87.90 & 23.89 \\
            CLIPArTT& 62.29 & 80.53 & 67.48 & 54.04 & 34.23 & 66.84 & 85.63 & 25.65 \\    
            TDA& 53.05 & 85.99 & 58.17 & 58.48 & 27.14 & 73.39 & 77.56 & 30.27 \\
            DPE& 64.00 & 86.38 & 57.42 & 58.80 & 37.35 & 74.01 & 77.01 & 30.66 \\
            \midrule
            \ouyangfan{CLIPTTA}& \ouyangfan{44.67} & \ouyangfan{56.87} & \ouyangfan{91.72} & \ouyangfan{32.44} & \ouyangfan{24.70} & \ouyangfan{30.86} & \ouyangfan{97.99} & \ouyangfan{8.85} \\
            STAMP& 60.19 & 79.13 & 70.28 & 51.85 & 32.06 & 65.95 & 86.63 & 23.87 \\
            UniEnt+EATA& 62.45 & 90.32 & 38.62 & 59.34 & 34.57 & 71.20 & 78.33 & 27.14 \\
            UniEnt+TPT& 65.00 & 82.18 & 70.76 & 56.59 & 35.11 & 36.20 & 93.50 & 24.08 \\
            \midrule
            \textbf{ProtoDCS} & \textbf{\ouyangfan{67.37 $\pm$ 0.53}} & \textbf{\ouyangfan{95.44 $\pm$ 0.02}} & \textbf{\ouyangfan{24.34 $\pm$ 0.07}} & \textbf{\ouyangfan{65.95 $\pm$ 0.47}} & \textbf{\ouyangfan{37.44 $\pm$ 0.20}} & \textbf{\ouyangfan{76.29 $\pm$ 0.02}} & \textbf{\ouyangfan{75.10 $\pm$ 0.01}} & \textbf{\ouyangfan{33.55 $\pm$ 0.15}} \\
            
            \bottomrule
    \end{tabular}
    
    
    \end{threeparttable}
    \vspace{-5mm}
\end{table*}

\subsection{Main Results}
\label{sec:results}
In this section, we validate the effectiveness of ProtoDCS on robust csID/csOOD separation and safe adaptation from three perspectives: state-of-the-art (SOTA) performance, architectural generality, and task extensibility.

\noindent\textbf{{State-of-the-art Performance}}.
ProtoDCS establishes a new SOTA on standard benchmarks, effectively alleviating the fundamental conflict between rejecting csOOD samples and adapting to csID data in OSTTA.
As detailed in Table~\ref{tab:cifar_benchmarks_c10_c100}, ProtoDCS outperforms all baselines across every metric on CIFAR-10-C and CIFAR-100-C. 
From comparison, we find the VLM-based TTA methods in the second group achieve systematically better performance than the standard TTA methods in the first group. 
This may be attributed to the involvement of the text modality in CLIP, which provides additional semantic information. 
However, directly extending existing OSTTA method to the CLIP framework (STAMP) or combining OSTTA method with existing TTA methods (UniEnt+EATA and UniEnt+TPT) in the third group fails to improve the TTA performance in the open-set scenario. 
For example, the Acc of STAMP (60.19\%) is generally lower than that of the VLM-based close-set methods, e.g., CLIPArTT (62.29\%) and DPE (64.00\%), on CIFAR-10-C. UniEnt+EATA and UniEnt+TPT solely attain marginal improvement compared with EATA and TPT on all metrics, respectively. 
This phenomenon indicates that existing OSTTA methods developed for small networks are not applicable to the highly compact and well-structured feature embedding space of VLMs. 
\luowei{Similarly, CLIPTTA exhibits marked performance degradation under the data-stream test setting (e.g., 44.67\% and 24.7\% on CIFAR-10-C and CIFAR-100-C, respectively), intrinsically due to its reliance on batch-level statistics for adaptation, rather than sample-wise adaptive feedback.}

These limitations highlight the necessity of a dedicated and robust separation mechanism for VLM-based TTA methods. Our ProtoDCS addresses this need via a two-stage verification mechanism and achieves significant performance gains across all metrics.
For example, ProtoDCS achieves respectively 95.59\% and 24.03\% on AUROC and FPR@TPR95 on CIFAR-10-C, a relative improvement over DPE by 10.69\% and 58.15\%, which signifies that the two-stage verification mechanism designed in ProtoDCS can effectively disentangle csID and csOOD representations within the highly compact and well-structured feature space of CLIP. 
In known-class prediction accuracy, ProtoDCS improves Acc to 67.98\% on CIFAR-10-C, an improvement over the best VLM-based closed-set TTA method DPE by relative 6.22\%, which reveals that the evidence-driven uncertainty-aware (EDUA) loss based residual optimization effectively prevents prototypes evolving from samples with high uncertainty, safeguarding adaptation on reliable csID samples to avoid the overconfidence pitfalls of entropy minimization.
As a result, these modules synergistically enhance adaptation to csID data while preserving discriminability against unknown classes, promoting respectively a 66.71\% and a 33.48\% OSCR on CIFAR-10-C and CIFAR-100-C, which comprehensively manifests the dual capability of ProtoDCS in “separation” and “adaptation” for OSTTA.


\begin{table}[!t]
    \centering
    \setlength{\tabcolsep}{2pt}
    
    \caption{Performance comparison using the CLIP ResNet-50 backbone, complementing the CLIP ViT-based results in Table~\ref{tab:cifar_benchmarks_c10_c100}. 
    }
    \label{tab:cifar_benchmarks_c10_c100_rn}
    \begin{threeparttable}
        
    \begin{tabular}{@{}lcccc@{}}
            \toprule
            \multicolumn{1}{c}{\multirow{2}{*}[-0.4ex]{Method}}
            & \multicolumn{4}{c}{CIFAR-10-C} \\
            
            \cmidrule(lr){2-5}

            & Acc$\uparrow$ & AUROC$\uparrow$ & FPR@TPR95$\downarrow$ & OSCR$\uparrow$ \\
            
            \midrule

            CLIP(RN)& 41.17 & 58.91 & 89.44 & 18.34 \\

            \midrule
            
            TENT& 21.40 & 61.94 & 88.44 & 14.44 \\
            SAR& 19.46 & 67.54 & 82.71 & 14.73  \\
            EATA& 18.64 & 57.60 & 91.17 & 12.11  \\

            \midrule
            
            TPT& 19.80 & 44.23 & 96.38 & 12.13  \\
            CLIPArTT & 21.38 & 62.35 & 87.79 & 14.44  \\
            TDA        & 25.17 & 70.84 & 83.91 & 30.77 \\
            DPE& 39.83 & 71.71 & 83.04 & 31.44 \\    

            \midrule  
            \ouyangfan{CLIPTTA} & \ouyangfan{12.22} & \ouyangfan{47.04} & \ouyangfan{93.99} & \ouyangfan{6.44} \\
            UniEnt+EATA& 18.61 & 57.94 & 91.14 & 12.06 \\
            UniEnt+TPT& 13.20 & 43.86 & 99.93 & 7.94 \\
            
            \midrule


        
            \textbf{ProtoDCS} & \textbf{\ouyangfan{43.95 $\pm$ 0.43}} & \textbf{\ouyangfan{77.19 $\pm$ 0.56}} & \textbf{\ouyangfan{82.35 $\pm$ 0.55}} & \textbf{\ouyangfan{37.20 $\pm$ 0.42}} \\
            
            \bottomrule
        \end{tabular}

    
        \end{threeparttable}
        \vspace{-3mm}
\end{table}

\noindent\textbf{Architectural generality}. ProtoDCS is ready to generalize to different backbone architectures and maintains its SOTA performance on all metrics. Table~\ref{tab:cifar_benchmarks_c10_c100_rn} shows the performance of ProtoDCS on CIFAR-10-C by using the CLIP ResNet-50 backbone. 
The results illustrate the same trend as those analyzed in Table~\ref{tab:cifar_benchmarks_c10_c100}, although they have a systematic performance degradation due to the limited capability of ResNet-50 compared with ViT. 
For example, UniEnt+EATA and UinEnt+TPT only attain comparable or even worse results compared with EATA and TPT on Acc (18.61\% vs. 18.64\% and 13.20\% vs. 19.80\%) and AUROC (57.94\% vs. 57.60\% and 43.86\% vs. 44.23\%), signifying that existing OSTTA methods developed for small networks are not ready to extend to VLM architectures. 
It further verifies the correctness of our claim that the feature embedding space of VLMs needs dedicated and flexible separation mechanism for robust separation. 
In contrast, ProtoDCS shows consistent improvements on all metrics with large gaps and achieves the SOTA performance, due to the intrinsic EDUA-based residual optimization for prototype update and the two-stage verification mechanism.
The results of ProtoDCS again validate the fact that the “separation” and “adaptation” capability of ProtoDCS for OSTTA can be inherently generalized across architectures.

\begin{table}[!t]
    \centering
    \setlength{\tabcolsep}{2pt}
    \caption{Results (\%) of different methods on Tiny-ImageNet-C using the CLIP ViT-B/16 backbone. $\uparrow$ indicates that larger values are better, and vice versa. 
    Bold indicates the best.}
    \label{tab:tiny}
    \begin{threeparttable}
    \begin{tabular}{lcccc}
        \toprule
            \multicolumn{1}{c}{\multirow{2}{*}[-0.4ex]{Method}}
            & \multicolumn{4}{c}{Tiny-ImageNet-C} \\
                
            \cmidrule(lr){2-5} & Acc$\uparrow$ & AUROC$\uparrow$ & FPR@TPR95$\downarrow$ & OSCR$\uparrow$ \\

            \midrule

            CLIP(ViT)  & 29.66 & 51.50 & 94.35 & 17.98 \\

            \midrule
            
            TENT & 28.92 & \textbf{62.50} & \textbf{88.42} & 20.51 \\

            SAR& 29.89 & 60.35 & 90.02 & 21.14\\

            EATA& 28.52 & 61.30 & 89.08 & 20.06\\

            \midrule

            TPT & 28.72 & 37.56 & 98.05 & 20.80 \\
            CLIPArTT & 29.25 & 62.27 & 88.57 & 20.65 \\
            TDA & 29.67 & 58.20 & 91.80 & 21.08 \\
            DPE  & 29.78 & 58.49 & 91.73 & 21.26 \\

            \midrule
            \ouyangfan{CLIPTTA} & \ouyangfan{25.43} & \ouyangfan{58.70} & \ouyangfan{91.21} & \ouyangfan{18.55} \\
            
            STAMP& 27.26 & 62.09 & 87.71 & 19.45 \\

            UniEnt+EATA & 29.54 & 61.36 & 88.86 & 20.91 \\
            
            UniEnt+TPT & 28.89 & 40.60 & 97.41 & 19.77 \\
            
            \midrule
            
            \ouyangfan{ProtoDCS} & \ouyangfan{\textbf{31.15 $\pm$ 0.17}} & \ouyangfan{58.04 $\pm$ 0.02} & \ouyangfan{92.08 $\pm$ 0.06} & \ouyangfan{\textbf{23.74 $\pm$ 0.12}}\\
            
            \bottomrule
        \end{tabular}
        \end{threeparttable}
        \vspace{-5mm}
\end{table}

\noindent\textbf{Task extensibility}. 
ProtoDCS demonstrates superior scalability on the more challenging Tiny-ImageNet-C dataset, where the visual distribution is significantly more complex with higher image resolutions, more classes, and pronounced similarity among classes. 
As shown in Table~\ref{tab:tiny}, 
while the VLM-based TTA methods are limited to exploiting the advantage of architectures compared with the standard TTA methods, ProtoDCS consistently brings 4.80\% and 11.67\% relative improvement over DPE on Acc and OSCR, respectively, with comparable separation ability on AUROC (58.32\% vs. 58.49\%).
This cross-task resilience is attributed to the intrinsic EDUA-based residual optimization for prototype update in the two-stage verification mechanism. 
In complex scenarios with 200 classes, the overlap between csID and csOOD distributions intensifies. 
Heuristic filtering (used in STAMP or UniEnt) becomes too brittle, misclassifying noise as informative samples. 
ProtoDCS, however, dynamically models this uncertainty, improving the filtration effects on ambiguous OOD samples that would otherwise corrupt the prototypes. 
This confirms that our approach generalizes well to large-scale, noise-prone environments where existing OSTTA methods falter.

\begin{table}[!t]
    \centering
    \small
    \setlength{\tabcolsep}{4pt} 
    \centering
    \caption{Ablation of Evidence-driven Uncertainty-aware Loss Components. The baseline (Row 1) uses entropy minimization.}
    \label{tab:ablation_loss}
    \begin{NiceTabular}{@{}c c c c c c c c@{}}[vlines={2,5}]
        \toprule
        \Block{2-1}{Method} & \Block{1-3}{\shortstack[c]{$\mathcal {L}$\textsubscript{uncertainty}}} &&& \Block{1-4}{CIFAR-100-C} \\
        \cmidrule(lr){2-4} \cmidrule(lr){5-8}
        & AU & EU & $\mathcal{L}_{\text{align}}$ & Acc$\uparrow$ & AUROC$\uparrow$ & FPR$\downarrow$ & OSCR$\uparrow$ \\
        \midrule
        \Block{5-1}{ProtoDCS} 
        & & & \checkmark & 36.42 & 75.31 & 75.21 & 32.47 \\
        & & \checkmark & \checkmark & 37.23 & 75.73 & 75.38 & 31.50 \\
        & \checkmark & & \checkmark & 36.42 & 76.29 & 75.21 & 32.49 \\
        & \checkmark & \checkmark & & 37.09 & 76.29 & 76.11 & 32.94 \\
        & \checkmark & \checkmark & \checkmark & \textbf{37.62} & \textbf{76.56} & \textbf{75.19} & \textbf{33.48} \\
        \bottomrule
    \end{NiceTabular}
    \vspace{-5mm}
\end{table}%

\subsection{Ablation Studies}
\label{sec:ablation}
We conduct comprehensive ablation studies on CIFAR-100-C using the CLIP ViT-B/16 backbone to validate the effectiveness of each component in ProtoDCS.

\noindent\textbf{C.1 Component Analysis.}
We dissect the contribution of our key modules through detailed ablation studies, separating the analysis into optimization objectives (Table~\ref{tab:ablation_loss}), and architectural components (Table~\ref{tab:ablation_components}).

\textbf{Entropy vs. Evidence:} 
As shown in Table~\ref{tab:ablation_loss}, standard entropy minimization (EM) is inherently flawed in open-set settings as it blindly amplifies prediction confidence regardless of correctness. 
In contrast, our proposed evidence-driven uncertainty-aware (EDUA) loss provides a principled solution by explicitly modeling both aleatoric and epistemic uncertainty, thereby preventing overconfidence and enabling better-calibrated prototype updates. 
Replacing the EDUA loss with EM (Row 1) or using partial uncertainty components (Rows 2-3) consistently harms performance across all metrics, particularly OSCR$\downarrow$. 
Furthermore, the alignment loss $\mathcal{L}_{\text{align}}$ proves essential for maintaining cross-modal semantic consistency; its removal (Row 4) leads to prototype drift and performance degradation, underscoring its role in preserving the joint embedding structure.

\begin{figure}[t]
    \centering
    \subfloat[]{
        \includegraphics[width=0.4\linewidth]{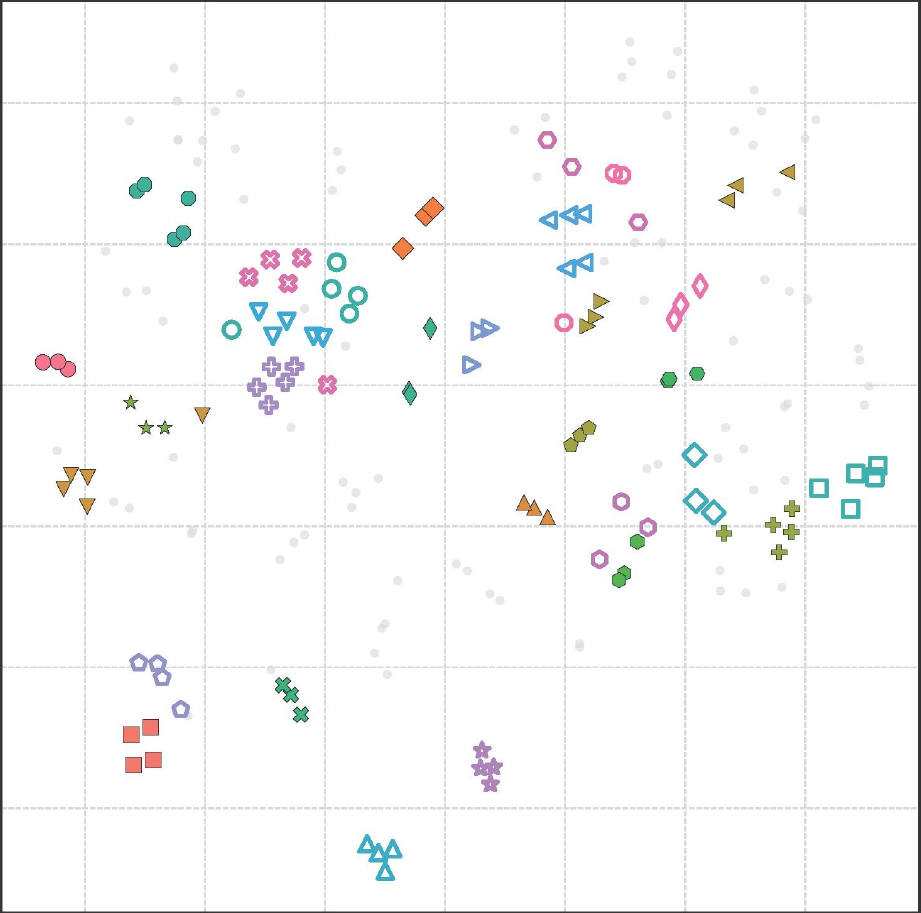}
    }
    \hspace{0.5cm}
    \subfloat[]{
        \includegraphics[width=0.4\linewidth]{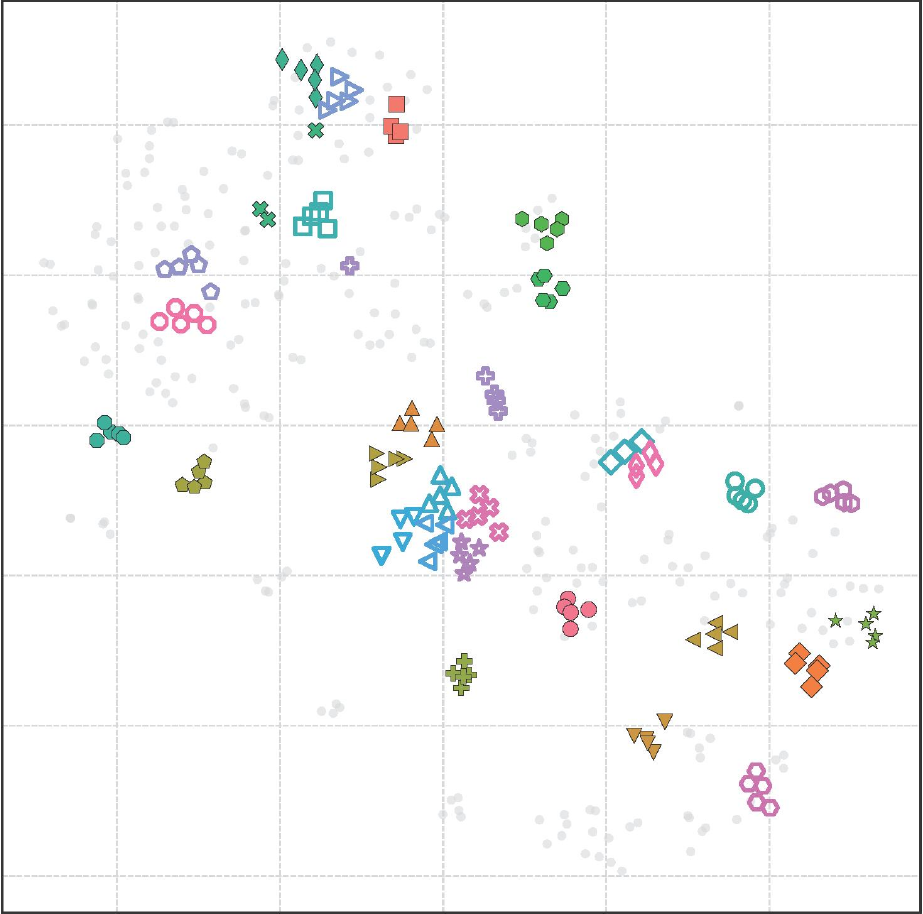}
    }
    \caption{
        \textbf{t-SNE Visualization of the Visual Cache on CIFAR-100-C.} 
        This figure illustrates how the cached features evolve over time, comparing the state after 3,000 samples (left) with the state after 30,000 samples (right). As more data is processed, our diversity-aware update mechanism creates increasingly compact and representative class clusters.
    }
    \vspace{-3mm}
    \label{supp_fig:tsne}
\end{figure}

\begin{table}[!t]
    \small
    \setlength{\tabcolsep}{2.7pt}
    \centering
    \caption{Ablation of Separation Mechanism and Prediction Strategy. We evaluate the impact of removing specific modules.}
    \label{tab:ablation_components}
    \begin{NiceTabular}{@{}c c c c c c c c c@{}}[vlines={2,6}]
        \toprule
        \Block{2-1}{Method} & \Block{1-4}{\shortstack[c]{Core Components}} &&&& \Block{1-4}{CIFAR-100-C} \\
        \cmidrule(lr){2-5} \cmidrule(lr){6-9}
        & Check & Verify & $\bm{P}_v$ & $\bm{P}_t$ & Acc$\uparrow$ & AUROC$\uparrow$ & FPR$\downarrow$ & OSCR$\uparrow$ \\
        \midrule
        \Block{5-1}{ProtoDCS} 
        & & \checkmark & \checkmark & \checkmark & 36.59 & 71.88 & 85.53 & 32.85 \\
        & \checkmark & & \checkmark & \checkmark & 37.15 & 76.33 & 75.20 & 33.00 \\
        & \checkmark & \checkmark & & \checkmark & 37.15 & 76.29 & 76.11 & 32.96 \\
        & \checkmark & \checkmark & \checkmark & & 37.09 & 76.29 & 77.11 & 32.94 \\
        & \checkmark & \checkmark & \checkmark & \checkmark & \textbf{37.62} & \textbf{76.56} & \textbf{75.19} & \textbf{33.48} \\
        \bottomrule
    \end{NiceTabular}
    \vspace{-5mm}
\end{table}
\textbf{Effectiveness of Double-Check:}
The first two rows of Table~\ref{tab:ablation_components} validate the critical, complementary roles of our two-stage verification design. 
Disabling the \textit{First-Check} (Row 1) causes a sharp decline in AUROC (71.88\% vs. 76.56\%), confirming its necessity for constructing a pure visual cache and preventing feature corruption from ambiguous samples. 
As visualized in Fig.~\ref{supp_fig:tsne}, this early filtering enables the formation of compact, discriminative class clusters in the cache. 
Conversely, relying solely on the First-Check without the GMM-based \textit{Final-Verification} (Row 2) fails to robustly handle distribution shifts, as static thresholds cannot adapt to the evolving test stream. The GMM, by dynamically modeling the decision boundary, provides the adaptive, probabilistic separation needed for ambiguous samples.

\textbf{Dual-Source Prediction:}
The last three rows of Table~\ref{tab:ablation_components} highlight the synergy of our dual-source prediction strategy. Relying exclusively on visual prototypes (Row 4) or textual prototypes (Row 3) yields inferior accuracy. The full approach (Row 5) effectively combines the semantic stability of textual prototypes with the distribution-aware adaptation of visual prototypes, confirming that multimodal fusion is vital for robust and accurate decision-making in open-set scenarios.

\noindent\textbf{C.2 Hyperparameter Sensitivity.}
We examine the stability of key hyperparameters on CIFAR-100-C in Tables~\ref{tab:sensitivity_gating} and \ref{tab:sensitivity_others} 
\luowei{and present more sensitivity analysis on Tiny-ImageNet-C to validate the robust generalizability of ProtoDCS across different datasets in SMs.}

\textbf{Gating and Cache Parameters ($\Theta_a, \Theta_b, \tau_{sim}$).} 
As shown in Table~\ref{tab:sensitivity_gating}, the performance is relatively robust to small variations in percentile thresholds, with $\Theta_a=Q(0.3)$ and $\Theta_b=Q(0.6)$ offering the optimal trade-off between cache purity and adaptation recall. 
This indicates that our dynamic percentiles effectively capture the relative confidence ranking within the sliding window $\mathcal{W}_c$, making it resilient to the absolute scale of openness scores. 
For the cache similarity threshold $\tau_{sim}$, a higher value ($\tau_{sim}=0.9$) yields slightly better accuracy (37.62\%) than lower values (e.g., $\tau_{sim}=0.6$, Acc 37.58\%). 
This confirms the benefit of maintaining high feature diversity within the cache, as it prevents redundancy and allows the visual prototypes to better represent the class manifold in the VLM's compact embedding space.

\begin{table}[!t]
\centering
\renewcommand{\arraystretch}{1.55} 
\setlength{\tabcolsep}{9pt}
\caption{Sensitivity analysis of Gating and Cache hyperparameters ($\Theta_a, \Theta_b, \tau_{sim}$) on CIFAR-100-C.}
\label{tab:sensitivity_gating}
\begin{tabular}{@{}c|lcccccc@{}}
    \hline
    \multirow{2}{*}{$\Theta_a$ } 
    & Value & 0.1 & 0.2 & \textbf{0.3} & 0.4 & 0.5 \\
    \cline{2-7}
    & Acc (\%) & 37.58 & 37.61 & \textbf{37.62}& 37.60 & 37.57 \\
    \hline
    \multirow{2}{*}{$\Theta_b$ } 
    & Value & 0.2 & 0.4 & \textbf{0.6} & 0.8 & 1.0 \\
    \cline{2-7}
    & Acc (\%) & 37.61 & 37.62 & \textbf{37.62} & 37.62 & 37.60 \\
    \hline
    \multirow{2}{*}{$\tau_{\text{sim}}$} 
    & Value & 0.6 & 0.7 & 0.8 & \textbf{0.9} & 1.0 \\
    \cline{2-7}
    & Acc (\%) & 37.58 & 37.58 & 37.60 & \textbf{37.62} & 37.60\\
    \hline
\end{tabular}
\vspace{-5mm}
\end{table}

\textbf{Verification, Window, and Loss Parameters.} 
The results in Table~\ref{tab:sensitivity_others} verify the robustness and design rationale of our framework across key hyperparameters and reveal \luowei{four} principal insights. 
1) The model's accuracy remains stable at 37.62\% for the GMM decision threshold $\Theta_p$ within [0.3, 0.5]. 
This plateau indicates that the GMM successfully learns a well-separated, bimodal distribution of openness scores for csID and csOOD samples, underscoring the robustness of the probabilistic separation mechanism.
2) Both the openness score window $\mathcal{W}c$ and the GMM fitting window $\mathcal{W}g$ achieve optimal performance with a size of 100 samples. This balanced setting provides sufficient data for reliable estimation while remaining promptly responsive to distribution shifts in the online stream. 
\luowei{3) The model is nearly insensitive to the initial fallback threshold $\Theta_c$ across [0.4, 0.9], with accuracy fluctuating within only $0.01\%$. This robustness stems from the GMM-based verification: once activated, it overrides the initial threshold with a data-driven boundary (around 0.745), effectively neutralizing the impact of $\Theta_c$ after the early stage.}
4) Performance is robust to the alignment weight $\lambda_{align}$ within a range of [0.2, 0.5], underscoring the stability of cross-modal consistency. Crucially, increasing the aleatoric uncertainty weight $\lambda_{AU}$ from 0.0 to 1.0 brings a significant accuracy improvement from 37.10\% to 37.62\%. 
This quantitatively demonstrates that explicitly modeling and leveraging data-inherent (aleatoric) uncertainty is as crucial as addressing model (epistemic) uncertainty for achieving robust and safe adaptation in noisy, open-set data streams.

\begin{table}[!t]
\centering
\renewcommand{\arraystretch}{1.55}
\setlength{\tabcolsep}{5pt}
\vspace{-3mm}
\caption{Sensitivity analysis of Verification ($\Theta_p$), Window Sizes ($\mathcal{W}_c, \mathcal{W}_g$), and Loss Weights ($\lambda_{align}, \lambda_{AU}$) on CIFAR-100-C.}
\label{tab:sensitivity_others}
\begin{tabular}{@{}c|lcccccc@{}}
    \hline
    \multirow{2}{*}{$\Theta_p$} 
    & Value & 0.3 & 0.4 & \textbf{0.5} & 0.6 & 0.7 & 0.9 \\
    \cline{2-8}
    & Acc (\%) & 37.62 & 37.62 & \textbf{37.62} & 37.61 & 37.60 & 37.56 \\
    \hline
    \multirow{2}{*}{$\mathcal{W}_g$} 
    & Value & 10 & 30 & 50 & \textbf{100} & 250 & 500 \\
    \cline{2-8}
    & Acc (\%) & 35.35 & 36.25 & 37.55 & \textbf{37.62} & 37.62 & 37.62 \\
    \hline
    \multirow{2}{*}{$\mathcal{W}_c$} 
    & Value  & 10 & 30 & 50 & \textbf{100} & 250 & 500 \\
    \cline{2-8}
    & Acc (\%) & 36.19 & 36.54 & 36.60 & \textbf{37.62} & 37.62 & 37.62\\
    \hline
    \multirow{2}{*}{\ouyangfan{$\Theta_c$}} 
    & \ouyangfan{Value} & \ouyangfan{0.4} & \ouyangfan{0.5} & \ouyangfan{0.6} & \textbf{\ouyangfan{0.7}} & \ouyangfan{0.8} & \ouyangfan{0.9} \\
    \cline{2-8}
    & \ouyangfan{Acc (\%)} & \ouyangfan{37.61} & \ouyangfan{37.61} & \ouyangfan{37.61} & \textbf{\ouyangfan{37.62}} & \ouyangfan{37.61} & \ouyangfan{37.61} \\
    \hline
    \multirow{2}{*}{$\lambda_{align}$} 
    & Value & 0.1 & \textbf{0.2} & 0.3 & 0.5 & 0.8 & 1 \\
    \cline{2-8}
    & Acc (\%) & 37.57 & \textbf{37.62} & 37.61 & 37.60 & 37.55 & 37.52 \\
    \hline
    \multirow{2}{*}{$\lambda_{AU}$} 
    & Value & 0.0 & 0.5 & \textbf{1} & 1.5 & 2 & 3 \\
    \cline{2-8}
    & Acc (\%) & 37.10 & 37.60 & \textbf{37.62} & 37.42 & 37.11 & 37.09 \\
    \hline
\end{tabular}
\vspace{-5mm}
\end{table}

\begin{figure*}[!t]
    \centering
    \subfloat[\small{Confidence Distribution}]{
        \includegraphics[width=0.31\textwidth]{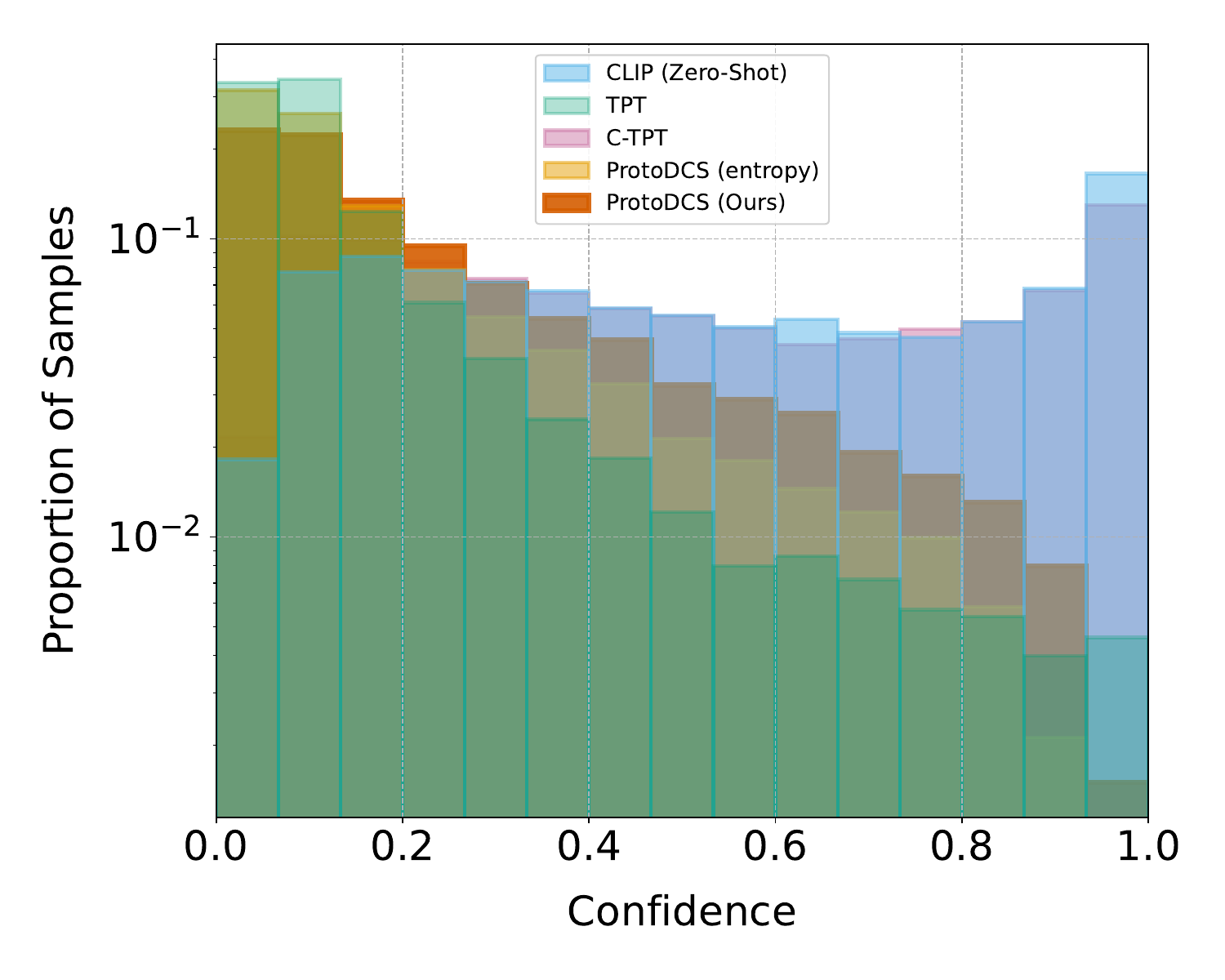}
        \label{fig:conf_dist}
    }
    \hfill
    \subfloat[\small{Reliability Diagram}]{
        \includegraphics[width=0.31\textwidth]{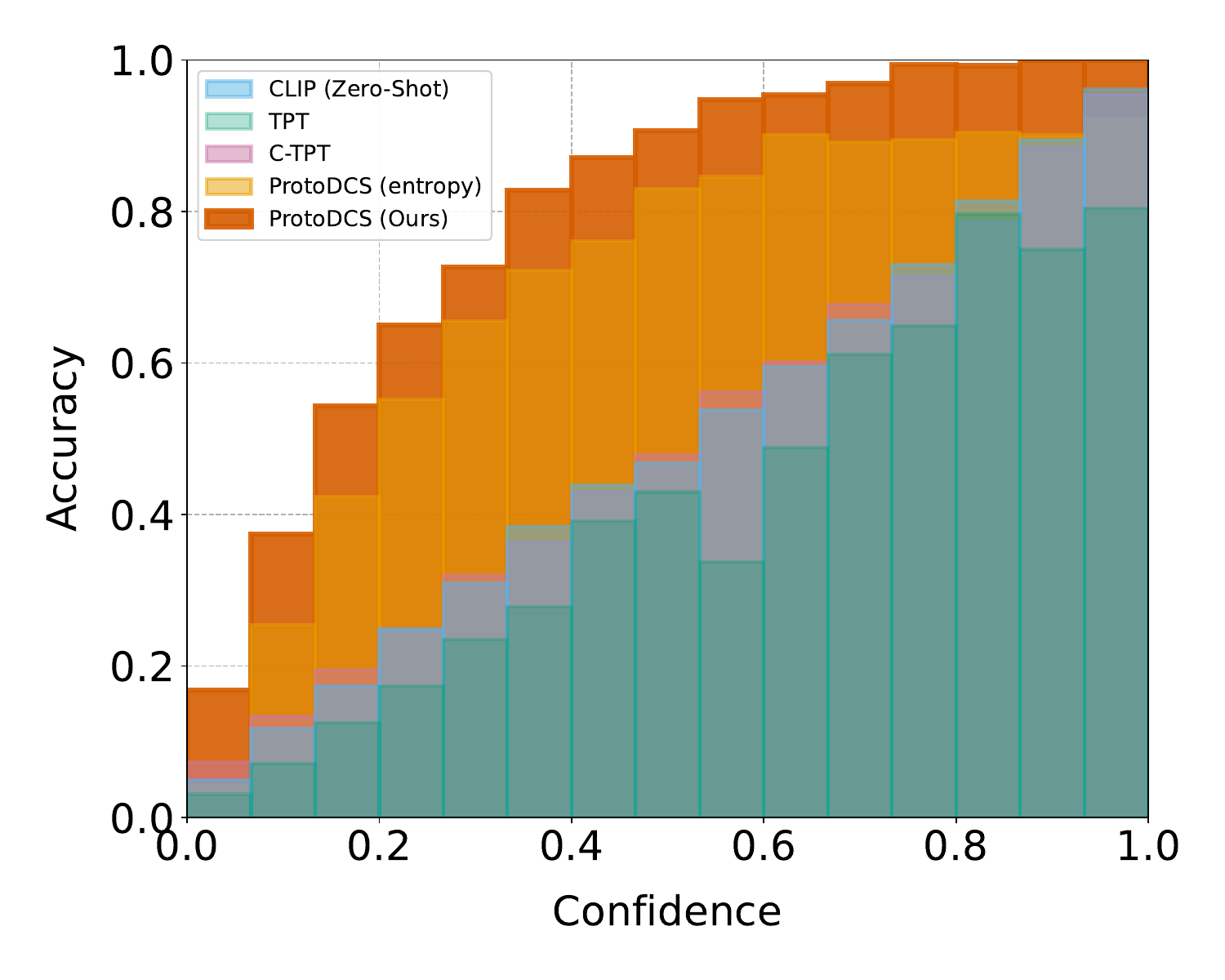}
        \label{fig:rel_diag}
    }
    \hfill
    \subfloat[\small{Accuracy-Uncertainty Curve}]{
        \includegraphics[width=0.31\textwidth]{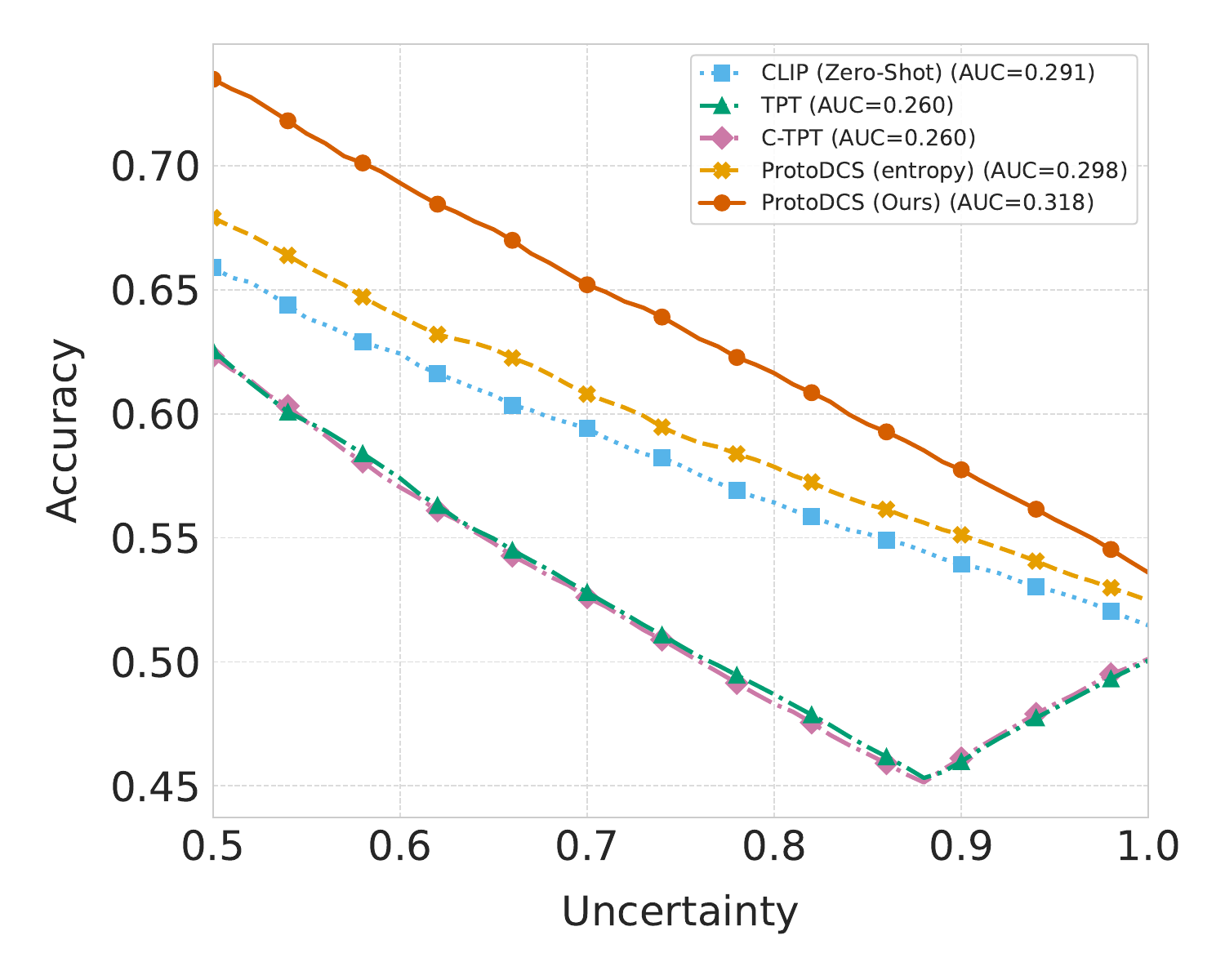}
        \label{fig:au_curve}
    }
    \caption{\textbf{Calibration and Uncertainty Analysis on Tiny-ImageNet-C.} This figure provides a tripartite analysis demonstrating the safe adaptation capability of ProtoDCS in open-set TTA. \luowei{All models are implemented in the same open-set TTA setting.} 
    (a) Confidence Distribution: TPT exhibits a skewed, low-confidence distribution, indicating predictive confusion induced by Entropy Minimization (EM). ProtoDCS recovers a balanced, well-calibrated distribution through its evidence-driven uncertainty-aware (EDUA) loss.
    (b) Reliability Diagram: While C-TPT shows perfect calibration (diagonal alignment), ProtoDCS adopts a deliberately conservative strategy (curve above diagonal), systematically assigning underconfidence on uncertain samples to enhance safety against OOD data.
    (c) Accuracy-Uncertainty Curve: ProtoDCS achieves the highest AUC, proving its superior ability to rank samples by risk (high uncertainty for OOD/hard ID), which is crucial for reliable sample filtering during adaptation. Replacing the EDUA loss with standard entropy (ProtoDCS(entropy)) yields a lower AUC.
    }
    \label{fig:calibration}
    \vspace{-3mm}
\end{figure*}

\begin{figure}[!t]
    \centering
    \includegraphics[width=\columnwidth]{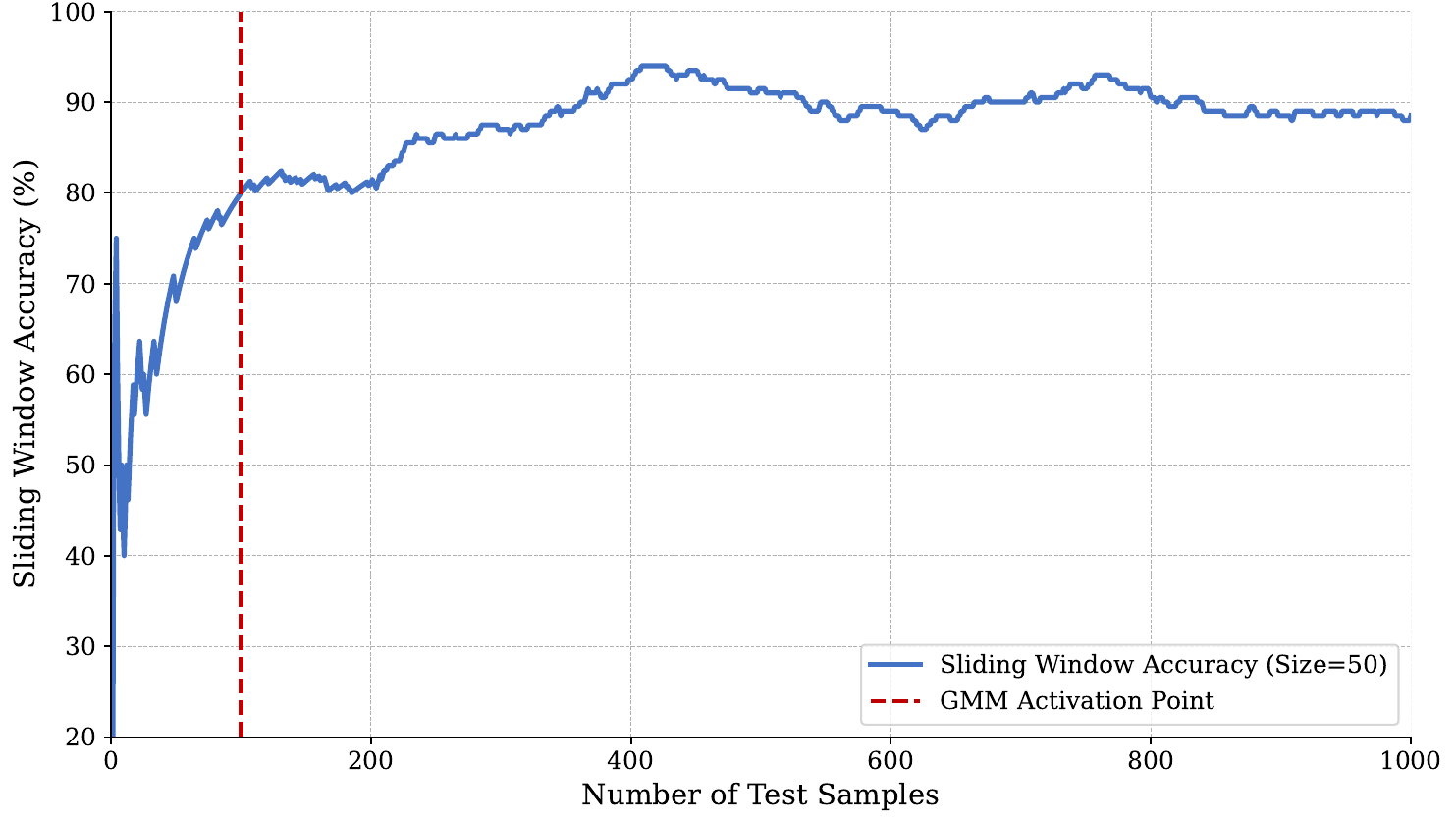}
    \caption{\textbf{Cold-Start Adaptation Dynamics on CIFAR-100-C.} The red dashed line marks the GMM activation point (100th sample). The curve illustrates the transition from initial fluctuation to rapid performance climbing, and finally to a robust steady state, confirming the effectiveness of our progressive double-check mechanism.}
    \vspace{-5mm}
    \label{fig:cold_start}
\end{figure}

\noindent\textbf{C.3 Cold-Start Adaptation Dynamics.}
To evaluate the online adaptation capability of ProtoDCS from scratch, we analyze the accuracy trajectory on the CIFAR-100-C test stream (Fig.~\ref{fig:cold_start}) \luowei{and defer the more in-depth cache purity analysis on skewed data streams in the cold-start stage to SMs.} 
The adaptation process exhibits three distinct phases:

\textbf{Phase 1: Initial Instability (0-100 samples).} 
Before the GMM activates ($n_{proc} < 100$), the model relies solely on heuristic percentile thresholds for sample filtering. The performance fluctuates significantly (50\%-60\%), revealing the inherent limitation of simple thresholding in handling ambiguous boundary samples. However, the overall upward trend indicates that our visual cache mechanism begins to accumulate useful features even in this early stage.
    
\textbf{Phase 2: Rapid Adaptation (100-400 samples).} 
A decisive turning point occurs at the 100th sample (red dashed line) when the GMM-based verification is activated. The accuracy curve immediately shifts to a steep and steady ascent, climbing rapidly from $\sim$80\% to over 90\%. This transformation validates the superiority of our probabilistic separation mechanism: by filtering high-quality csID samples, it guides the lightweight prototype updates onto a correct and efficient trajectory.
    
\textbf{Phase 3: Steady State (400+ samples).} After processing approximately 400 samples, the model converges to a plateau, stabilizing between 88\% and 94\%. The curve fluctuates narrowly within a high-level corridor rather than degrading, demonstrating the robustness of our closed-loop system against continuous noise and distribution shifts in the open-set stream.

\noindent\textbf{\ouyangfan{C.4 Robustness under Skewed Test Streams.}} 
\luowei{To assess the GMM bimodal assumption in the context of extreme class imbalance, we systematically alter the ratio of csID to csOOD samples, ranging from 5\%:95\% to 95\%:5\%. 
Our observations indicate that although extreme skewness slightly reduces GMM convergence due to the scarcity of the minority component, our fallback mechanism averts threshold collapse, thereby ensuring stable open-set classification. 
We refer readers to SMs for detailed analyses of the Hellinger distance, convergence statistics, and robust boundary evaluations.}

\subsection{Safe Adaptation} 
Fig.~\ref{fig:calibration} forms a coherent evidential chain that diagnoses the fundamental flaw of Entropy Minimization (EM) in open-set TTA and demonstrates how our Evidence-Driven Uncertainty-Aware (EDUA) loss enables safe adaptation in ProtoDCS.

Fig.~\ref{fig:conf_dist} (Confidence Distribution) reveals the ``predictive confusion'' induced by EM in open-set streams. Contrary to its tendency to cause overconfidence in closed-set settings, EM leads to a heavily left-skewed, low-confidence distribution for TPT. This is not a strategic shift towards caution but a manifestation of failed optimization: EM cannot coherently sharpen predictions for OOD and ambiguous boundary samples within VLMs' compact feature space, resulting in inconsistent, low-confidence outputs. ProtoDCS(entropy) alleviates this skew primarily through its two-stage filtering mechanism. ProtoDCS, however, addresses the root cause via the EDUA loss, which enables stratified, evidence-based confidence assignment. It successfully discriminates between OOD (low confidence), ambiguous ID (moderate confidence), and clear ID samples (appropriately high confidence), producing a balanced distribution that aligns with well-calibrated baselines like CLIP and C-TPT.

\begin{table*}[!t]
\centering
\setlength{\tabcolsep}{12pt}
\caption{Performance comparisons on ImageNet and its variants under a \textbf{Closed-Set} setting (Natural Distribution Shifts). \textbf{bold} and \underline{underline} indicate the best and second-best results, respectively.}
\label{tab:natural_shifts}
\begin{tabular}{lcccccc}
    \toprule
    Method & ImageNet & ImageNet-A & ImageNet-V2 & ImageNet-R & ImageNet-S & Average \\
    \midrule
    CLIP-ViT-B/16 & 66.73 & 47.87 & 60.86 & 73.98 & 46.09 & 59.11 \\
    \midrule
    TPT~\cite{shu2022TPT} & 68.98 & 54.77 & 63.45 & 77.06 & 47.94 & 62.44 \\
    C-TPT~\cite{ctpt@24iclr} & 69.30 & 52.90 & 63.40 & 78.00 & 48.50 & 62.42 \\
    DiffTPT~\cite{difftpt} & 70.30 & 55.68 & \underline{65.10} & 75.00 & 46.80 & 62.28 \\
    TDA~\cite{karmanov2024efficient} & 69.51 & 60.11 & 64.67 & 80.24 & 50.54 & 65.01 \\
    TPS~\cite{niu2023SAR} & 70.19 & 60.08 & 64.73 & 80.27 & 49.95 & 65.04 \\
    \luowei{BCA}~\cite{bayesianTTA25CVPR} & \luowei{70.22} & \luowei{\textbf{61.14}} & \luowei{64.90} & \luowei{\textbf{80.72}} & \luowei{50.87} & \luowei{65.37} \\
    DPE~\cite{zhang2024DPE} & \textbf{71.91} & 59.63 & \textbf{65.44} & 80.40 & \textbf{52.26} & \textbf{65.93} \\
    \midrule
    \textbf{ProtoDCS (Ours)} & \underline{71.07} & \underline{60.24} & 65.01 & \underline{80.69} & \underline{51.43} & \underline{65.69} \\
    \bottomrule
\end{tabular}
\vspace{-5mm}
\end{table*}

\begin{table}[!t]
    \centering
    \setlength{\tabcolsep}{13pt}
    \caption{Efficiency evaluation on Tiny-ImageNet-C with ViT-B/16 backbone. We report Throughput (samples/sec $\uparrow$), Latency (ms/sample $\downarrow$), and Peak GPU Memory (MB $\downarrow$). \textbf{bold} and \underline{underlined} indicate the best and second-best results \textit{among adaptation methods}, respectively.}
    \label{tab:efficiency}
    \begin{tabular}{lccc}
        \toprule
            \multicolumn{1}{c}{\multirow{2}{*}[-0.4ex]{Method}}
            & \multicolumn{3}{c}{Efficiency Metrics} \\
            \cmidrule(lr){2-4} & Throughput$\uparrow$ & Latency$\downarrow$ & Memory$\downarrow$ \\
            
            \midrule
            CLIP (ViT) & 172.48 & 5.80 & 356 \\
            \midrule
            TENT & 45.66 & 21.90 & 12471 \\
            TPT & 2.49 & 401.78 & 3795 \\
            DPE  & \textbf{80.93} & \textbf{12.40} & \textbf{356} \\
            
            \midrule
            \textbf{ProtoDCS} & \underline{55.86} & \underline{17.90} & \underline{372}\\
            \bottomrule
            
        \end{tabular}
        \vspace{-5mm}
\end{table}

Fig.~\ref{fig:rel_diag} (Reliability Diagram) evaluates the ``safety'' of this calibration from a risk perspective. TPT's curve below the diagonal confirms its overconfidence. While C-TPT achieves excellent calibration (diagonal alignment), such perfect confidence-accuracy agreement can be misleading in OSTTA, as it may not reflect high confidence on OOD samples. ProtoDCS adopts a deliberately conservative strategy, with its curve lying predominantly above the diagonal. This indicates a systematic tendency to assign lower confidence than the empirical accuracy -- a safety-oriented bias that inherently mitigates the risk of making high-confidence errors on unfamiliar or ambiguous data.

Fig~\ref{fig:au_curve} (Accuracy-Uncertainty Curve) quantifies the practical ``decision-making utility'' of each method's uncertainty estimate. The Area Under this Curve (AUC) measures how well the model's uncertainty ranks samples by their true risk (e.g., being OOD or hard ID). The low and identical AUC for TPT and C-TPT (0.260) proves that entropy alone is a poor proxy for risk in open-set scenarios. ProtoDCS(entropy) shows improvement (AUC 0.298), attributable to its filtering mechanism. ProtoDCS achieves the highest AUC (0.320), providing direct evidence for the superiority of the EDUA loss. By explicitly modeling and distinguishing epistemic (for OOD) and aleatoric (for boundary ID) uncertainty, EDUA provides a more accurate risk-ranking signal. This allows ProtoDCS to effectively identify and reject high-risk samples, ensuring that adaptation is driven primarily by low-uncertainty, high-fidelity csID data.

In summary, the three subfigures in Fig.~\ref{fig:calibration} demonstrates that EDUA moves beyond merely avoiding overconfidence. It establishes a principled, evidence-based framework for uncertainty quantification that is crucial for safe, reliable, and effective adaptation in open-world environments.

\subsection{Efficiency Evaluation}
To assess the feasibility for online deployment, we evaluate the computational efficiency on Tiny-ImageNet-C using a workstation equipped with an AMD EPYC 7T83 CPU and a single NVIDIA RTX 4090 GPU (24GB VRAM). As summarized in Table~\ref{tab:efficiency}, ProtoDCS achieves an optimal Pareto frontier, balancing minimal resource consumption with robust open-set performance.
Existing methods face severe bottlenecks: TENT incurs a massive memory footprint (12,471 MB) due to full-model gradient tracking, while TPT suffers from prohibitive latency (2.49 samples/sec). In sharp contrast, ProtoDCS maintains a high throughput of 55.86 samples/sec and requires only 372 MB of peak memory—a marginal 16 MB increase over the inference-only Zero-shot CLIP baseline.
This efficiency stems from our architectural design: by operating exclusively at the prototype level, ProtoDCS bypasses the costly backpropagation through the heavy ViT backbone. 
DPE, a dual-stream prototype evolution method designed for closed-set TTA, achieves the best raw efficiency metrics. This highlights the inherent advantage of prototype-level adaptation. 
However, DPE lacks any mechanism for csID/csOOD separation, which fundamentally limits its performance in open-set scenarios (see Table.~\ref{tab:cifar_benchmarks_c10_c100}). 
ProtoDCS is explicitly architected for OSTTA. Its robust two-stage verification mechanism introduces only a minimal computational overhead, while its core optimization component -- the EDUA-based residual optimization -- is both mathematically concise and computationally lightweight. 
Consequently, ProtoDCS offers a practical solution for edge devices where memory is scarce, delivering sophisticated uncertainty estimation and safety without the latency overhead typical of test-time optimization methods. See SMs for the complexity analysis in theory.

\vspace{-1mm}
\subsection{Performance on Closed-set TTA}
\label{sec:further_results}
A natural concern for open-set TTA methods is whether the sample separation mechanism inadvertently discards \textit{hard} in-distribution samples, thereby degrading performance in closed-set scenarios where no true OOD samples exist. 
To address this, we evaluate ProtoDCS on the standard ImageNet benchmark and its four naturally shifted variants: ImageNet-A/V2/R/S. 
As presented in Table~\ref{tab:natural_shifts}, ProtoDCS demonstrates remarkable robustness even when operating under its full separation framework designed for open-set streams. It achieves an average accuracy of 65.69\%, which is highly comparable to the state-of-the-art closed-set specialist DPE (65.93\%).
\luowei{The 0.2\% performance drop stems from the adoption of a different prediction function in both, rather than from misclassifying csID samples as csOOD. See SMs for detailed experiments and analysis.} 
Crucially, on datasets containing samples with severe visual deviations (e.g., ImageNet-A/R), ProtoDCS matches or even outperforms DPE (e.g., 60.24\% vs. 59.63\% on ImageNet-A and 80.69\% vs. 80.40\% on ImageNet-R), \luowei{and it also outperforms BCA~\cite{bayesianTTA25CVPR} on ImageNet-V2/S, achieving a superior average precision.}
These results confirm that our two-stage verification mechanism is sufficiently discriminative to preserve challenging in-distribution samples for adaptation while effectively filtering out only irrelevant or ambiguous signals, thereby validating its safety and broad applicability across diverse data regimes -- from pure closed-set adaptation to complex open-set environments.

\section{Conclusion}
\label{sec:conclusion}

\noindent In this paper, we introduce ProtoDCS, a robust framework for VLM-based open-set TTA. ProtoDCS \revise{mitigates the brittleness of rigid thresholding} with a probabilistic double-check separation mechanism, reliably filtering out-of-distribution samples prior to adaptation. For safe adaptation, we abandon overconfidence-prone entropy minimization in favor of an evidence-driven uncertainty-aware loss, which explicitly models aleatoric and epistemic uncertainty to preserve model calibration. For efficient adaptation, ProtoDCS operates exclusively at the prototype level with a frozen backbone, eliminating costly gradient propagation through the VLM. Experiments demonstrate that ProtoDCS achieves state-of-the-art performance in both known-class accuracy and OOD detection, while maintaining high throughput and low memory footprint. This work enables the safe, efficient, and robust deployment of VLMs in dynamic open-world environments. 
\luowei{Theoretically, ProtoDCS may suffer from semantic drift during long-term adaptation to fine-grained test streams, as subtle inter-class visual differences are easily confounded, thereby contaminating the visual cache -- a risk we will address in future work.}

\vfill
\appendices
\newpage

\section*{Supplementary Materials}


\markboth{Journal of \LaTeX\ Class Files,~Vol.~14, No.~8, August~2021}%
{Shell \MakeLowercase{\textit{et al.}}: A Sample Article Using IEEEtran.cls for IEEE Journals}



\maketitle

\renewcommand{\thefigure}{S\arabic{figure}}
\renewcommand{\thetable}{S\arabic{table}}
\renewcommand{\theequation}{S\arabic{equation}}
\setcounter{figure}{0}
\setcounter{table}{0}
\setcounter{equation}{0}

We organize our supplementary materials as follows:

\begin{itemize}
    \item \textbf{Appendix~\ref{sec:alg}} details the overall algorithm of ProtoDCS.

    \item \textbf{Appendix~\ref{sec:exp}} provides more experimental setups.

    \item \textbf{Appendix~\ref{sec:ana}} provides a detailed breakdown of the diversity-aware visual cache, including its update policy and a computational complexity analysis.
    
\end{itemize}

\appendices

\section{OVERALL ALGORITHM} \label{sec:alg}

This section details the foundational components of the ProtoDCS approach. We begin by presenting the complete online adaptation process in the form of pseudocode (Algorithm~\ref{alg:protodcs}). This algorithm provides a step-by-step overview of the pipeline. 

\subsection{OVERALL ALGORITHM OF PROTODCS}
\label{app:pseudocode}
Algorithm~\ref{alg:protodcs} outlines the complete online adaptation process of ProtoDCS for a single incoming test sample. It details the sequential execution of the three core stages outlined in the main paper: openness-aware initialization, evidence-driven optimization, and final verification, which together form our double-check separation and prototype evolution pipeline.

\textit{Firstly}, in the initial check stage (lines 3-10), we compute an initial openness score $S_{\text{open}}$ and use it to conditionally update the diversity-aware visual cache $\mathcal{C}_v$.
\textit{Secondly}, based on a stricter trust threshold, we perform a conditional, evidence-driven adaptation (lines 11-19). This stage generates temporary prototypes $\bm{P}'_t$ and $\bm{P}'_v$ by optimizing small, learnable residuals, while keeping the main VLM backbone frozen.
\textit{Finally}, in the verification and finalization stage (lines 20-29), the sample is re-evaluated using these temporary prototypes. A Gaussian Mixture Model (GMM) makes the definitive csID/csOOD decision. If the sample is confirmed as csID, it is used for the final prediction and to evolve the global prototypes.

\begin{algorithm*}[h!]
\caption{ProtoDCS: Online Adaptation Process}
\label{alg:protodcs}
\begin{algorithmic}[1]
    \STATE \textbf{Initialize:} VLM visual encoder $E_v$, Global text prototypes $\bm{P}_t$, visual cache $\mathcal{C}_v \leftarrow \emptyset$, score windows $W_c, W_g \leftarrow \emptyset$, learning rates $\eta_t, \eta_v$, alignment weight $\lambda_{align}$, \textbf{GMM threshold $\Theta_p$, \textbf{fallback threshold $\Theta_c \leftarrow 0.7$}, sample counter $n \leftarrow 0$, processed counter $n_{proc} \leftarrow 0$, \textbf{quality threshold $\Theta_q \leftarrow 0.1$}, momentum $\gamma \leftarrow 0.01$}
    
    \STATE \textbf{for} each incoming sample $x$ \textbf{do}
    
    \STATE \quad \textbf{$n_{proc} \leftarrow n_{proc} + 1$} \textcolor{gray}{// Increment processed sample counter}
    \STATE
    \STATE \quad \textcolor{gray}{\# --- Stage 1: First-Check and Cache Update ---}
    
    \STATE \quad $\bm{f}_v \leftarrow E_v(x)$
    
    \STATE \quad $S_{\text{open}} \leftarrow 1 - \max_{k} \cos(\bm{f}_v, \bm{f}_{t,k})$ \quad \textcolor{gray}{// Compute initial openness score $S_{\text{open}}$ using Eq.~(\ref{eq:os_init})}
    
    \STATE \quad $\mathcal{W}_c \leftarrow \text{UpdateWindow}(\mathcal{W}_c, S_{\text{open}})$ \quad \textcolor{gray}{// Performs a FIFO (First-In, First-Out) update on the 100-element score window}
    
    \STATE \quad Set $\Theta_a$ to the 30th and $\Theta_b$ to the 60th percentile of scores in $\mathcal{W}_c$ \quad \textcolor{gray}{// Set cache and trust thresholds}

    \STATE \quad \textbf{if} $S_{\text{open}} < \Theta_a$ \textbf{then}
    
        \STATE \quad \quad Update visual cache $\mathcal{C}_v$ with $(x, \bm{f}_v)$ \quad \textcolor{gray}{// See diversity-aware logic in Sec.~(\ref{sec:initial_assessment}) and Algorithm~\ref{alg:cache_update}}
        
    \STATE \quad \textbf{end if}
    \STATE
    \STATE \quad \textcolor{gray}{\# --- Stage 2: Conditional Evidence-driven Adaptation ---}
    \STATE \quad $\bm{P}_v \leftarrow \text{AverageFeatures}(\mathcal{C}_v)$ \quad \textcolor{gray}{// Compute visual prototypes}
    \STATE \quad \textbf{if} $S_{\text{open}} < \Theta_b$ \textbf{then} \quad \textcolor{gray}{// Trustworthy sample for adaptation}
        
        \STATE \quad \quad $\Delta \bm{P}_t, \Delta \bm{P}_v \leftarrow \bm{0}$ \quad \textcolor{gray}{// Initialize residuals}
        
        \STATE \quad \quad $\mathcal{L} \leftarrow \mathcal{L}_{\text{uncertainty}} + \lambda_{\text{align}} \mathcal{L}_{\text{align}}$ \quad \textcolor{gray}{// From Eq.~(\ref{eq:total_loss_condensed_revised})}
        
        \STATE \quad \quad Update $\Delta \bm{P}_t, \Delta \bm{P}_v$ with one step of gradient descent: $\nabla \mathcal{L}$
        \STATE \quad \quad$\bm{P}'_t, \bm{P}'_v \leftarrow \text{Norm}(\bm{P}_t + \Delta \bm{P}_t), \text{Norm}(\bm{P}_v + \Delta \bm{P}_v)$ \quad \textcolor{gray}{// using Eq.~(\ref{eq:tempo_protos})}
    \STATE \quad \textbf{else}
        \STATE \quad \quad$\bm{P}'_t, \bm{P}'_v \leftarrow \bm{P}_t, \bm{P}_v$ \quad \textcolor{gray}{// No adaptation}
    \STATE \quad \textbf{end if}
    \STATE
    
    \STATE \quad \textcolor{gray}{\# --- Stage 3: Final Verification with GMM ---}
    \STATE \quad $S'_{\text{open}} \leftarrow 1 - \max_{k} \cos(\bm{f}_v, \bm{p}'_{t,k})$ \quad \textcolor{gray}{// Re-evaluate with temporary prototypes}
    
    \STATE \quad $\mathcal{W}_g \leftarrow \text{UpdateWindow}(\mathcal{W}_g, S'_{\text{open}})$

    \STATE \quad \textbf{if} $n_{proc} \ge 100$ \textbf{then} \quad \textcolor{gray}{// Activate GMM only after collecting enough samples}

    \STATE \quad \quad $\mathcal{G}, \ouyangfan{\mu_1, \mu_2} \leftarrow \text{FitGMM}(\mathcal{W}_g)$ \quad \textcolor{gray}{// Fit GMM and extract component means}

    \STATE \quad \quad \textbf{$P(C_{\text{csOOD}}|S'_{open}) \leftarrow \text{GetPosterior}(\mathcal{G}, S'_{open})$} \quad \textcolor{gray}{// Update decision with GMM posterior}

    \STATE \quad \quad \ouyangfan{$\Theta_c \leftarrow (\mu_1 + \mu_2) / 2$} \quad \textcolor{gray}{\ouyangfan{// Update fallback threshold with GMM midpoint}}
    
    \STATE \quad \textbf{else}

    \STATE \quad \quad \textbf{$P(C_{\text{csOOD}}|S'_{\text{open}}) \leftarrow \mathbbm{1}[S'_{\text{open}} \ge \Theta_c]$} \quad \textcolor{gray}{// Determine using a fixed score threshold}
    
    \STATE \quad \textbf{end if}

    \STATE
    \STATE \quad \textcolor{gray}{\# --- Stage 4: Conditional Prediction and Asymmetric Global Evolution ---}
    \STATE \quad \textbf{if} $P(c_{\text{csOOD}} | S'_{\text{open}}) < \Theta_p$ \textbf{then} \quad \textcolor{gray}{// Verify using Eq.~(\ref{eq:gmm_ood_classification_revised})}
        \STATE \quad \quad Predict with $\bm{P}'_t, \bm{P}'_v$ using Eq.~(\ref{eq:l_final_computation})(\ref{eq:affinity})
        \STATE \quad \quad Calculate $\mathcal{AU}(x)$ using Eq.~(\ref{eq:au})
        
        \STATE \quad \quad \textbf{if} $\mathcal{AU}(x) < \Theta_q$ \textbf{then} \quad \textcolor{gray}{// See Appendix.~(\ref{sec:ana}) for details}
        
            \STATE \quad \quad \quad \textbf{$n \leftarrow n + 1$} \quad \textcolor{gray}{// Increment confirmed csID sample counter}
            
            \STATE \quad \quad \quad $\bm{P}_t \leftarrow (1-1/n)\bm{P}_t + (1/n)\bm{P}'_t$ \quad \textcolor{gray}{// Evolve global Text prototypes $\bm{P}_t$ using Eq.~(\ref{eq:cma_update_revised})}

            \STATE \quad \quad \quad $\Theta_q \leftarrow (1-\gamma)\Theta_q + \gamma \cdot \mathcal{AU}(x)$ \quad \textcolor{gray}{// Update quality threshold via EMA}
            
        \STATE \quad \quad \textbf{end if}
            
        \STATE \quad \quad \textcolor{gray}{// Note: Global Visual Prototypes $\bm{P}_v$ evolve implicitly via cache updates in Stage 1}
    \STATE \quad \textbf{else}
        \STATE \quad \quad Classify $x$ as OOD
    \STATE \quad \textbf{end if}
\STATE \textbf{end for}
\end{algorithmic}
\end{algorithm*}




\section{More Experimental Setups} \label{sec:exp}
This section provides the compared methods and implementation details for the final-verification stage (Section IV-C) and the prototype evolution (Section IV-D).

\subsection{Compared methods}
We compare ProtoDCS against three groups of methods. \textbf{1) Standard TTA Methods:} These methods focus on closed-set adaptation. \textit{TENT}\cite{wang2021tent} minimizes prediction entropy by updating batch normalization parameters. \textit{SAR}\cite{niu2023SAR} employs sharpness-aware minimization to filter out noisy samples with high gradients. \textit{EATA}\cite{eata@22icml} combines reliable sample selection based on prediction consistency with Fisher information-based weight regularization.
\textbf{2) VLM-based TTA Methods:} These methods are closed-set methods but leverage the CLIP architecture. 
\textit{TPT}~\cite{shu2022TPT} adapts the model by tuning learnable textual prompts via entropy minimization. 
\textit{C-TPT}~\cite{ctpt@24iclr} adds a dispersion loss to TPT to calibrate predictions. 
\textit{DiffTPT}~\cite{difftpt} incorporates a diffusion model to generate diverse views for robust prompt tuning.        
\textit{CLIPArTT}~\cite{clipartt} updates the visual normalization layers via a transductive loss derived from instance-specific multi-class prompts. 
\luowei{\textit{BCA}~\cite{bayesianTTA25CVPR} updates both the class likelihood and class prior for adaptation.}
\textit{TDA}~\cite{karmanov2024efficient} proposes a training-free dynamic adapter that maintains positive and negative key-value caches to refine predictions. 
\textit{DPE}~\cite{zhang2024DPE} introduces a dual-stream evolution strategy updating both textual and visual prototypes. 
\textbf{3) Open-Set TTA Methods:} These are explicitly designed for csOOD scenarios.
\textit{UniEnt}~\cite{guan2024UniEnt} proposes a unified optimization objective that filters csOOD samples via similarity with fixed text prototypes.
\textit{STAMP}~\cite{yu2024stamp} utilizes an outlier-aware memory replay mechanism to filter csOOD samples.
\luowei{\textit{CLIPTTA}~\cite{cliptta25nips} learns an adaptive threshold to gate the probability of being a csOOD sample.}
To eliminate the effects of backbone differences, we extend the open-set TTA (OSTTA) techniques originally designed for ResNet to CLIP for a fair comparison. Specifically, we extend UniEnt to TPT and STAMP to CLIP.
Compared with OSTTA methods on the same backbone, our gains could demonstrate that they stem from closed-loop systematic design rather than simple aggregation.
In addition, \textit{Zero-shot CLIP} serves as the fundamental lower bound. While our extensive evaluations focus on the CLIP-ViT backbone, we also validate ProtoDCS on CLIP-ResNet-50 backbone to verify its generalizability across different architectures.

\subsection{GMM-BASED SEPARATION LOGIC}
The GMM-based separation forms the core of our final verification step.
\begin{itemize}
    \item \textbf{Dynamic Fitting:} We do not fit a GMM for every sample. Instead, we maintain a sliding window of the most recent final openness scores, $S'_{open}$, and re-fit the two-component GMM on this window periodically (e.g., every 100 samples). This is both efficient and adaptive to non-stationary data streams.
    \item \textbf{Decision and Fallback:} After a successful GMM fit, we associate the component with the higher mean with the OOD component. A sample's posterior probability of belonging to this component determines its final classification. If a GMM fit fails to converge, our system robustly falls back to using the hard threshold $\Theta_c$, which was derived from the last successful GMM fit. This ensures a reliable decision is always made, even during periods of unstable data distribution.
\end{itemize}

\ouyangfan{Specifically, under extremely skewed test streams (e.g., highly imbalanced ratios of csID to csOOD samples), the bimodal distribution of openness scores may become less distinct due to the scarcity of the minority component, potentially leading to singularity in covariance estimation during GMM fitting. To handle such boundary cases, ProtoDCS incorporates a robust fallback mechanism (Algorithm~\ref{alg:protodcs}, Line 33). When the GMM expectation-maximization (EM) optimization converges successfully on the sliding window, we dynamically update the fallback threshold as the midpoint of the two component means:
\begin{equation}
\label{eq:fallback_midpoint}
\Theta_c \leftarrow \frac{\mu_1 + \mu_2}{2}.
\end{equation}
If GMM EM optimization occasionally fails to converge due to severe class imbalance, the system automatically falls back to this static threshold derived from the last successful GMM fit, ensuring stable decision boundaries and preventing threshold instability.}

\begin{figure}[h]
    \centering
    \includegraphics[width=0.6\linewidth]{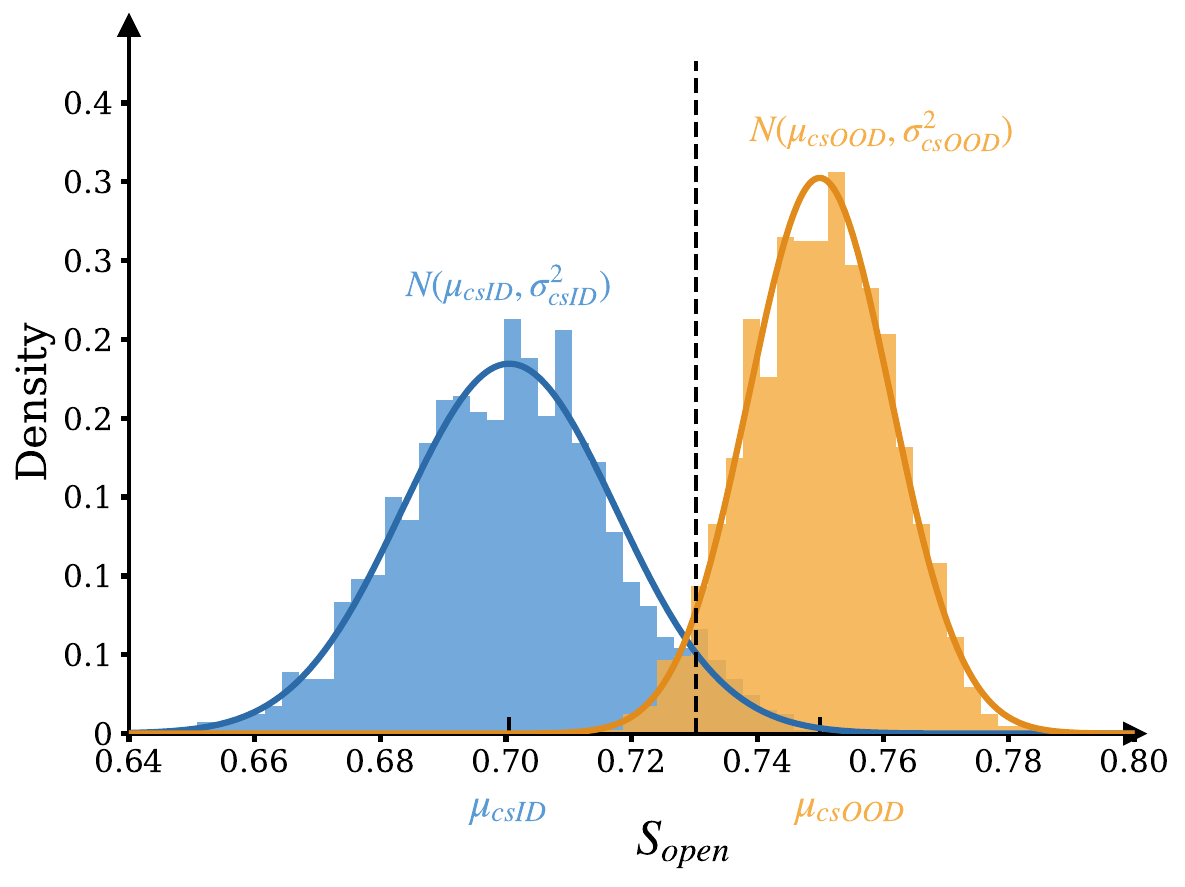}
    \caption{
        \textbf{Distribution of Final Openness Scores ($S'_{\text{open}}$) on CIFAR-10-C.} 
        This plot shows the score distribution for ID and OOD samples under the ``brightness'' corruption. The clear separation demonstrates the effectiveness of our temporary prototype refinement in enhancing the score's discriminability.}
    \label{supp_fig:gmm}
\end{figure}

Fig.~\ref{supp_fig:gmm} illustrates the resulting distribution of the final openness scores, showing a clear separation between ID and OOD samples that the GMM is designed to model.

\subsection{\ouyangfan{GMM ROBUSTNESS UNDER SKEWED TEST STREAMS}}
\begin{figure}[htbp]
    \centering
    \includegraphics[width= 1\linewidth]{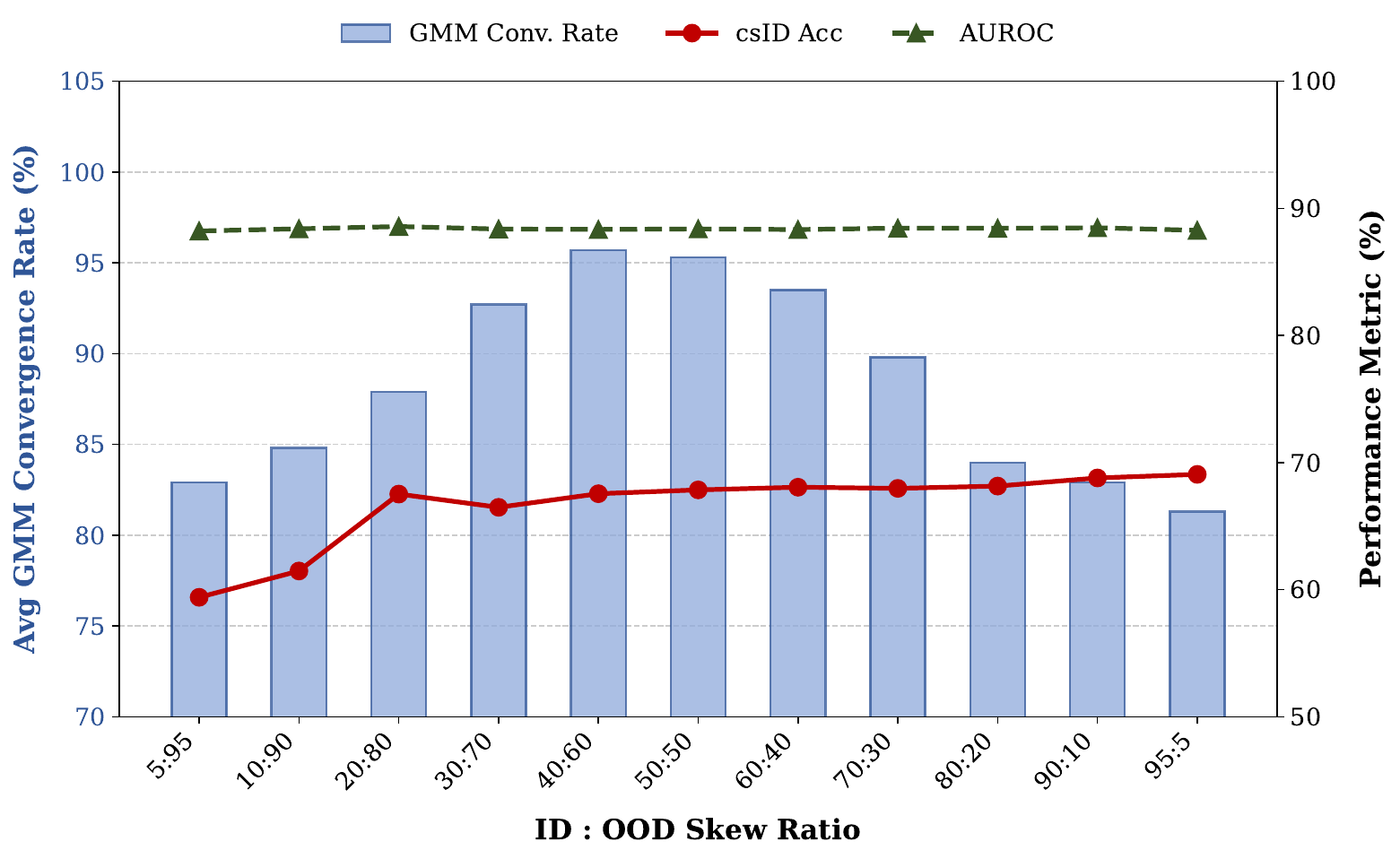}
    \caption{\ouyangfan{Evaluation of ProtoDCS under varying csID:csOOD skew ratios on CIFAR-10-C. The background bar chart represents GMM convergence rate (left Y-axis), while the curves represent csID Acc and AUROC (right Y-axis).}}
    \label{fig:skew_robustness_chart_sm}
\end{figure}

\ouyangfan{To evaluate the generalizability and robustness of the GMM bimodal assumption in ProtoDCS under extreme class imbalance, we systematically vary the ratio of csID to csOOD samples from 5:95 to 95:5 while maintaining a constant test stream length of 10,000 for fairness. We employ the squared Hellinger Distance~\cite{hellinger1909neue} to quantitatively measure the separation between the fitted Gaussian components $\mathcal{N}(\mu_1, \sigma_1^2)$ and $\mathcal{N}(\mu_2, \sigma_2^2)$:
\begin{equation}
H^2(P, Q) = 1 - \sqrt{\frac{2\sigma_1\sigma_2}{\sigma_1^2 + \sigma_2^2}} \exp\left(-\frac{(\mu_1 - \mu_2)^2}{4(\sigma_1^2 + \sigma_2^2)}\right),
\label{eq:hellinger}
\end{equation}
where $H^2(P,Q) \in [0, 1]$, and a value closer to 1 indicates higher component separation. As presented in Table~\ref{tab:skew_robustness_sm} and visualized in Fig.~\ref{fig:skew_robustness_chart_sm}, we observe that under balanced streams (50:50), GMM expectation-maximization (EM) optimization achieves its highest convergence rate (95.3) and maximum component separation (Avg H-Dist of 0.6777) because both components have equal sample sizes ($N=50$) within the 100-sample sliding window, ensuring a full-rank covariance matrix. Under extremely skewed ratios (e.g., 5:95), the GMM EM convergence rate marginally degrades to approximately 81.3 because the minority component has too few samples ($N \le 5$), causing covariance singularity. However, our fallback mechanism seamlessly takes over to prevent threshold collapse. Consequently, csID accuracy monotonically scales from 59.40 to 69.07 without negative transfer, while AUROC remains highly stable as a flat line hovering around 88.3, demonstrating robust open-set classification under extreme domain drifts.}

\begin{table}[htbp]
    \centering
    \setlength{\tabcolsep}{9pt}
    \caption{\ouyangfan{Evaluation of ProtoDCS across varying csID:csOOD skew ratios on CIFAR-10-C.$\uparrow$ indicates that larger values are better, and vice versa. All values are percentages (\%). Bold indicates the best.}}
    \label{tab:skew_robustness_sm}
    \begin{tabular}{ccccc}
    \toprule
    \ouyangfan{ID:OOD} & \ouyangfan{GMM Conv$\uparrow$} & \ouyangfan{H-Dist$\uparrow$} & \ouyangfan{Acc$\uparrow$} & \ouyangfan{AUROC$\uparrow$} \\
    \midrule
    \ouyangfan{5:95}  & \ouyangfan{82.9} & \ouyangfan{0.4151} & \ouyangfan{59.40} & \ouyangfan{88.22} \\
    \ouyangfan{10:90} & \ouyangfan{84.8} & \ouyangfan{0.5085} & \ouyangfan{61.47} & \ouyangfan{88.39} \\
    \ouyangfan{20:80} & \ouyangfan{87.9} & \ouyangfan{0.6490} & \ouyangfan{67.52} & \textbf{\ouyangfan{88.57}} \\
    \ouyangfan{30:70} & \ouyangfan{92.7} & \ouyangfan{0.6447} & \ouyangfan{66.48} & \ouyangfan{88.37} \\
    \ouyangfan{40:60} & \textbf{\ouyangfan{95.7}} & \ouyangfan{0.6748} & \ouyangfan{67.55} & \ouyangfan{88.34} \\
    \ouyangfan{50:50} & \ouyangfan{95.3} & \textbf{\ouyangfan{0.6777}} & \ouyangfan{67.85} & \ouyangfan{88.38} \\
    \ouyangfan{60:40} & \ouyangfan{93.5} & \ouyangfan{0.6677} & \ouyangfan{68.06} & \ouyangfan{88.33} \\
    \ouyangfan{70:30} & \ouyangfan{89.8} & \ouyangfan{0.6428} & \ouyangfan{67.97} & \ouyangfan{88.44} \\
    \ouyangfan{80:20} & \ouyangfan{84.0} & \ouyangfan{0.6524} & \ouyangfan{68.15} & \ouyangfan{88.43} \\
    \ouyangfan{90:10} & \ouyangfan{82.9} & \ouyangfan{0.5726} & \ouyangfan{68.79} & \ouyangfan{88.47} \\
    \ouyangfan{95:5}  & \ouyangfan{81.3} & \ouyangfan{0.5046} & \textbf{\ouyangfan{69.07}} & \ouyangfan{88.27} \\
    \bottomrule
    \end{tabular}
\end{table}

\begin{table*}[htbp]
    \centering
    \setlength{\tabcolsep}{15pt}
    \caption{\ouyangfan{Quantitative evaluation of standard Softmax Entropy and Epistemic Uncertainty (EU) for correct in-distribution (csID) samples versus misclassified out-of-distribution (csOOD) samples. Standard deviation offsets ($\sigma$-distance) relative to correct ID samples are reported. ``Intercepted'' indicates that these pseudo-high-confidence samples are successfully intercepted by our EDUA-guided separation mechanism.}}
    \label{tab:edua_discrimination_sm}
    \begin{tabular}{llcccc}
    \toprule
    \ouyangfan{Dataset} & \ouyangfan{Sample Group} & \ouyangfan{Avg Entropy} & \ouyangfan{Entropy Offset} & \ouyangfan{Avg EU} & \ouyangfan{EU Offset} \\
    \midrule
    \ouyangfan{CIFAR-10-C} 
    & \ouyangfan{Correct csID}     & \ouyangfan{1.3698} & \ouyangfan{-}  & \ouyangfan{0.6959} & \ouyangfan{-} \\
    & \ouyangfan{Misclassified csOOD} & \ouyangfan{1.2193} & \ouyangfan{-0.26$\sigma$} & \ouyangfan{0.7740} & \ouyangfan{+0.96$\sigma$ (Intercepted)} \\
    \midrule
    \ouyangfan{CIFAR-100-C} 
    & \ouyangfan{Correct csID}     & \ouyangfan{3.2515} & \ouyangfan{-}  & \ouyangfan{0.7351} & \ouyangfan{-} \\
    & \ouyangfan{Misclassified csOOD} & \ouyangfan{2.9335} & \ouyangfan{-0.32$\sigma$} & \ouyangfan{0.7563} & \ouyangfan{+0.48$\sigma$ (Intercepted)} \\
    \midrule
    \ouyangfan{Tiny-ImageNet-C} 
    & \ouyangfan{Correct csID}     & \ouyangfan{3.7229} & \ouyangfan{-}  & \ouyangfan{0.7040} & \ouyangfan{-} \\
    & \ouyangfan{Misclassified csOOD} & \ouyangfan{3.0644} & \ouyangfan{-0.57$\sigma$} & \ouyangfan{0.7054} & \ouyangfan{+0.04$\sigma$ (Intercepted)} \\
    \bottomrule
    \end{tabular}
\end{table*}

\subsection{\ouyangfan{CLOSED-SET REJECTION-RATE AND PERFORMANCE GAP ANALYSIS}}
\ouyangfan{To investigate the marginal performance gap (approximately 0.30\%) between ProtoDCS and DPE in pure closed-set scenarios, we first evaluate the false OOD rejection rate of our separation mechanism. This analysis is conducted across all five standard ImageNet benchmarks under pure closed-set settings where no true OOD samples are present. The quantitative results are summarized in Table S3:}

\begin{table}[htbp]
    \centering
    \setlength{\tabcolsep}{3pt}
    \caption{\ouyangfan{False OOD rejection-rate analysis of the ProtoDCS separation mechanism in pure closed-set scenarios.}}
    \label{tab:rebuttal_rejection_rate_sm}
    \begin{tabular}{lccc}
    \toprule
    \ouyangfan{Dataset} & \ouyangfan{Total Samples} & \ouyangfan{Rejected Samples} & \ouyangfan{Rejection Rate (\%)} \\
    \midrule
    \ouyangfan{ImageNet} & \ouyangfan{50,000} & \ouyangfan{0} & \ouyangfan{0.00} \\
    \ouyangfan{ImageNet-A} & \ouyangfan{7,500} & \ouyangfan{0} & \ouyangfan{0.00} \\
    \ouyangfan{ImageNet-R} & \ouyangfan{30,000} & \ouyangfan{0} & \ouyangfan{0.00} \\
    \ouyangfan{ImageNet-S} & \ouyangfan{50,000} & \ouyangfan{0} & \ouyangfan{0.00} \\
    \ouyangfan{ImageNet-V2} & \ouyangfan{10,000} & \ouyangfan{0} & \ouyangfan{0.00} \\
    \bottomrule
    \end{tabular}
\end{table}

\ouyangfan{As shown in Table S3, the false OOD rejection rate is strictly 0.00 across all closed-set datasets. Because the pure closed-set stream lacks OOD components, the GMM cannot fit a bimodal distribution of openness scores, which causes our system to fallback to the dynamic threshold $\Theta_c = 0.7$. Since the openness scores of ID samples are significantly lower than $\Theta_c$, exactly zero samples are misidentified as OOD. This solid evidence demonstrates that our separation mechanism does not inadvertently reject hard-to-classify csID samples, and the cost of safety at the separation stage is zero.}

\ouyangfan{We further analyze the mathematical formulation of the multimodal information fusion during the final prediction stage. Let $T_k$ denote the textual similarity logits and $V_k$ denote the visual affinity scores. The performance discrepancy is governed by the spatial design of the fusion formulas:}
\begin{enumerate}
    \item \ouyangfan{Logit-Level Unbounded Fusion (DPE): DPE directly adds the visual affinity to the raw textual logits prior to normalization: $\arg\max (T_k + V_k)$. Because the scale of textual logits $T_k$ is natively large ($T_k \approx 25$) and visual affinity is small ($V_k \approx 0.5$), $V_k$ merely acts as a 2.0\% perturbation in the raw logit space. This text-dominated behavior shields the prediction from visual cache impurities, marginally maximizing accuracy in zero-noise closed-set scenarios.}
    \item \ouyangfan{Probability-Space Bounded Fusion (ProtoDCS): ProtoDCS first projects the textual logits into a normalized probability space, and subsequently incorporates the visual feature affinity: $\arg\max (\text{Softmax}(T_k) + V_k)$. Here, the textual probability is strictly bounded in $[0, 1]$, while $V_k \in [0, 0.5]$. This formulation mathematically amplifies the relative weight of the visual term to over 35\%. While this provides exceptional capability to correct textual predictions using visual features, in a pure closed-set environment, even minor visual cache variances can introduce a 0.5\% level of interference to the textual predictions.}
\end{enumerate}

\ouyangfan{To empirically prove this mathematical causality, we conducted a rigorous ablation study where we temporarily replaced our probability-space fusion equation with DPE's logit-level unbounded fusion (Configuration B). The results across multiple random seeds and all five datasets are summarized in Table S4:}

\begin{table*}[htbp]
    \centering
    \setlength{\tabcolsep}{10pt}
    \caption{\ouyangfan{Ablation study on the multimodal prediction fusion formulas in pure closed-set scenarios (Acc $\uparrow$).}}
    \label{tab:rebuttal_fusion_ablation_sm}
    \begin{tabular}{lcccccc}
    \toprule
    \ouyangfan{Method} & \ouyangfan{ImageNet} & \ouyangfan{ImageNet-A} & \ouyangfan{ImageNet-V2} & \ouyangfan{ImageNet-R} & \ouyangfan{ImageNet-S} & \ouyangfan{Average} \\
    \midrule
    \ouyangfan{DPE} & \ouyangfan{71.95} & \ouyangfan{59.62} & \ouyangfan{65.58} & \ouyangfan{80.44} & \ouyangfan{52.28} & \ouyangfan{65.97} \\
    \ouyangfan{ProtoDCS (Original)} & \ouyangfan{71.04} & \ouyangfan{60.23} & \ouyangfan{65.01} & \ouyangfan{80.63} & \ouyangfan{51.43} & \ouyangfan{65.67} \\
    \ouyangfan{ProtoDCS (Modified)} & \ouyangfan{71.93} & \ouyangfan{59.61} & \ouyangfan{65.60} & \ouyangfan{80.43} & \ouyangfan{52.25} & \ouyangfan{65.96} \\
    \bottomrule
    \end{tabular}
\end{table*}

\ouyangfan{As demonstrated in Table S4, once the bounded fusion in ProtoDCS is replaced with DPE's logit-level direct addition, the average accuracy of ProtoDCS perfectly regresses and equals the performance level of DPE (65.96 vs 65.97). This confirms that the minor performance gap is entirely attributable to the mathematical constraint of the fusion mechanism, rather than the GMM separation module. In an open-set environment, if logit-level unbounded fusion is employed, any minor visual feature perturbation from an unseen OOD sample will be exponentially amplified during the subsequent Softmax calculation, leading to severe representation degradation. By constraining the textual output to the probability space and introducing the visual feature as a linearly bounded residual, ProtoDCS strictly limits the maximum mathematical interference an OOD visual feature can impose on the global prediction distribution, ensuring stable open-set adaptation.}

\subsection{CONDITIONAL AND ASYMMETRIC PROTOTYPE EVOLUTION}
\label{app:conditional_evolution}
After the final csID/csOOD decision, the evolution of the global prototypes ($\bm{P}_t$ and $\bm{P}_v$) is highly conditional and asymmetric, ensuring robustness.

For confirmed csID samples, the update is gated by a \textit{dynamic quality threshold}. Only csID samples whose final prediction yields a high confidence are used for evolution. In our implementation, this confidence is measured by the sample's final predictive uncertainty (aleatoric uncertainty), where a lower uncertainty value indicates higher confidence. To achieve a threshold that \textbf{tightens as adaptation proceeds}, we initialize it with a static base value 0.1 and then update it using an Exponential Moving Average (EMA). This threshold tightens as adaptation proceeds, ensuring that only increasingly reliable samples contribute.
\begin{itemize}
\item \textit{Textual Prototype Update:} For a high-quality csID sample, the global text prototypes $\bm{P}_t$ are updated slowly via a Cumulative Moving Average (CMA) with the temporary prototypes $\bm{P}'_t$. This smooths out noise from individual samples and captures persistent semantic shifts.
\item \textit{Visual Prototype Update:} The visual feature of a high-quality csID sample is first placed in a temporary buffer. This buffer is periodically flushed, and its most representative samples are then added to the visual cache $\mathcal{C}_v$ using the diversity-aware logic from Algorithm~\ref{alg:cache_update}. This delayed, batched update for the visual cache further enhances stability.
\end{itemize}
This conditional and asymmetric evolution mechanism allows ProtoDCS to adapt to csID data while actively reinforcing its ability to reject csOOD data.

\subsection{\ouyangfan{EVIDENCE-DRIVEN UNCERTAINTY DISCRIMINABILITY ANALYSIS}}
\ouyangfan{To clarify the mathematical robustness of our Evidence-Driven Uncertainty-Aware (EDUA) loss under incorrect, pseudo-high-confidence predictions on csOOD samples, we analyze the decoupling of absolute evidence from relative probability. Standard Softmax is translation-invariant, making it prone to relative probability inflation (low entropy/pseudo-high-confidence) even when absolute logit magnitudes are universally extremely low. Conversely, our EDUA models absolute Dirichlet evidence $S = \sum_{k=1}^K \alpha_k$ (where $\alpha_k = \text{ReLU}(z_k) + 1$). Since csOOD samples suffer from poor multimodal alignment with known prototypes, their absolute logits $z_k$ remain close to zero, forcing $S$ to be extremely small and yielding a very high epistemic uncertainty ($EU = K/S$).}

\ouyangfan{In the compact embedding space of CLIP, the calculated epistemic uncertainty (EU) of correct in-distribution (csID) samples exhibits an extremely narrow variance ($\sigma \approx 0.008$). In statistics, the significance of a distribution offset is evaluated by its standard deviation multiplier ($\sigma$-distance) rather than absolute numerical differences. We compare the average Softmax Entropy and Epistemic Uncertainty (EU) of misclassified (pseudo-high-confidence) csOOD samples against correct in-distribution (csID) samples across three benchmarks, as presented in Table~\ref{tab:edua_discrimination_sm}:}

\begin{table}[htbp]
    \centering
    \renewcommand{\arraystretch}{1.55}
    \setlength{\tabcolsep}{8pt}
    \caption{\ouyangfan{Sensitivity analysis of Gating and Cache hyperparameters ($\Theta_a, \Theta_b, \tau_{sim}$) on Tiny-ImageNet-C.}}
    \label{tab:sensitivity_gating_tiny_sm}
    \begin{tabular}{@{}c|lcccccc@{}}
        \hline
        \multirow{2}{*}{\ouyangfan{$\Theta_a$}} 
        & \ouyangfan{Value} & \ouyangfan{0.1} & \ouyangfan{0.2} & \textbf{\ouyangfan{0.3}} & \ouyangfan{0.4} & \ouyangfan{0.5} \\
        \cline{2-7}
        & \ouyangfan{Acc (\%)} & \ouyangfan{30.05} & \ouyangfan{31.25} & \ouyangfan{31.21} & \ouyangfan{30.02} & \ouyangfan{27.85} \\
        \hline
        \multirow{2}{*}{\ouyangfan{$\Theta_b$}} 
        & \ouyangfan{Value} & \ouyangfan{0.2} & \ouyangfan{0.4} & \textbf{\ouyangfan{0.6}} & \ouyangfan{0.8} & \ouyangfan{1.0} \\
        \cline{2-7}
        & \ouyangfan{Acc (\%)} & \ouyangfan{29.08} & \ouyangfan{31.18} & \ouyangfan{31.21} & \ouyangfan{31.02} & \ouyangfan{30.75} \\
        \hline
        \multirow{2}{*}{\ouyangfan{$\tau_{sim}$}} 
        & \ouyangfan{Value} & \ouyangfan{0.6} & \ouyangfan{0.7} & \ouyangfan{0.8} & \textbf{\ouyangfan{0.9}} & \ouyangfan{1.0} \\
        \cline{2-7}
        & \ouyangfan{Acc (\%)} & \ouyangfan{31.02} & \ouyangfan{31.15} & \ouyangfan{31.21} & \ouyangfan{31.21} & \ouyangfan{31.05} \\
        \hline
    \end{tabular}
\end{table}

\begin{figure*}[htbp]
    \centering
    \begin{tabular}{cc}
        \begin{tabular}{c}
            \includegraphics[width=0.2\linewidth]{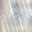} \\
            \ouyangfan{(a) Snow Image (CIFAR-10-C)} \\
            \ouyangfan{Entropy: 1.0746 ($-0.27\sigma$)} \\
            \ouyangfan{EU: 0.7270 ($+0.56\sigma$)}
        \end{tabular} &
        \begin{tabular}{c}
            \includegraphics[width=0.2\linewidth]{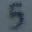} \\
            \ouyangfan{(b) Glass Blur Image 1 (CIFAR-100-C)} \\
            \ouyangfan{Entropy: 1.6221 ($-2.21\sigma$)} \\
            \ouyangfan{EU: 0.7810 ($+1.00\sigma$)}
        \end{tabular} \\
        \vspace{0.01em} \\ 
        \begin{tabular}{c}
            \includegraphics[width=0.2\linewidth]{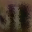} \\
            \ouyangfan{(c) Glass Blur Image 2 (CIFAR-100-C)} \\
            \ouyangfan{Entropy: 1.5244 ($-2.32\sigma$)} \\
            \ouyangfan{EU: 0.7704 ($+0.72\sigma$)}
        \end{tabular} &
        \begin{tabular}{c}
            \includegraphics[width=0.2\linewidth]{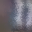} \\
            \ouyangfan{(d) Glass Blur Image 3 (CIFAR-100-C)} \\
            \ouyangfan{Entropy: 1.9008 ($-1.88\sigma$)} \\
            \ouyangfan{EU: 0.7851 ($+1.10\sigma$)}
        \end{tabular}
    \end{tabular}
    \caption{\ouyangfan{Four representative incorrectly classified (pseudo-high-confidence) csOOD samples that exhibit pseudo-low values under standard Softmax entropy and their corresponding uncertainty indicators. Standard Softmax entropy and Epistemic Uncertainty (EU) are reported, along with their respective deviations relative to correct in-distribution samples ($\sigma$-distance).}}
    \label{fig:pseudo_confidence_cases_sm}
\end{figure*}

\ouyangfan{As presented in Table~\ref{tab:edua_discrimination_sm}, the average entropy of misclassified csOOD samples is lower than that of correct in-distribution samples, yielding negative offsets (e.g., $-0.26\sigma$ on CIFAR-10-C and $-0.57\sigma$ on Tiny-ImageNet-C), demonstrating that standard Softmax falsely deems them to be highly confident. Conversely, our EU successfully maps these pseudo-high-confidence OOD samples to positive statistical offsets (e.g., up to $+0.96\sigma$ on CIFAR-10-C), enabling robust and reliable separation. To visually demonstrate this discriminability, we present four representative incorrectly classified csOOD samples (which exhibit pseudo-low values under standard Softmax entropy) along with their statistical indicators in Fig.~\ref{fig:pseudo_confidence_cases_sm}:}

\subsection{\ouyangfan{CROSS-DATASET HYPERPARAMETER GENERALIZATION}}

\ouyangfan{To address potential concerns regarding whether the chosen default hyperparameters overfit to a specific data distribution, we demonstrate the cross-dataset generalization capability of ProtoDCS by replicating the entire hyperparameter sensitivity analysis on the more complex Tiny-ImageNet-C dataset (averaged across 15 corruptions). We strictly maintain the same search grids as those used for the CIFAR datasets. The comprehensive evaluation results are presented in Table~\ref{tab:sensitivity_gating_tiny_sm} and Table~\ref{tab:sensitivity_others_tiny_sm}.}

\begin{table}[htbp]
    \centering
    \renewcommand{\arraystretch}{1.55}
    \setlength{\tabcolsep}{5pt}
    \caption{\ouyangfan{Sensitivity analysis of Verification ($\Theta_p$), Window Sizes ($\mathcal{W}_c, \mathcal{W}_g$), and Loss Weights ($\lambda_{align}, \lambda_{AU}$) on Tiny-ImageNet-C.}}
    \label{tab:sensitivity_others_tiny_sm}
    \begin{tabular}{@{}c|lcccccc@{}}
        \hline
        \multirow{2}{*}{\ouyangfan{$\Theta_p$}} 
        & \ouyangfan{Value} & \ouyangfan{0.3} & \ouyangfan{0.4} & \textbf{\ouyangfan{0.5}} & \ouyangfan{0.6} & \ouyangfan{0.7} & \ouyangfan{0.9} \\
        \cline{2-8}
        & \ouyangfan{Acc (\%)} & \ouyangfan{29.18} & \ouyangfan{30.22} & \ouyangfan{31.21} & \ouyangfan{31.15} & \ouyangfan{30.95} & \ouyangfan{30.65} \\
        \hline
        \multirow{2}{*}{\ouyangfan{$\mathcal{W}_g$}} 
        & \ouyangfan{Value} & \ouyangfan{10} & \ouyangfan{30} & \ouyangfan{50} & \textbf{\ouyangfan{100}} & \ouyangfan{250} & \ouyangfan{500} \\
        \cline{2-8}
        & \ouyangfan{Acc (\%)} & \ouyangfan{29.50} & \ouyangfan{30.80} & \ouyangfan{31.12} & \ouyangfan{31.21} & \ouyangfan{31.15} & \ouyangfan{31.02} \\
        \hline
        \multirow{2}{*}{\ouyangfan{$\mathcal{W}_c$}} 
        & \ouyangfan{Value}  & \ouyangfan{10} & \ouyangfan{30} & \ouyangfan{50} & \textbf{\ouyangfan{100}} & \ouyangfan{250} & \ouyangfan{500} \\
        \cline{2-8}
        & \ouyangfan{Acc (\%)} & \ouyangfan{30.12} & \ouyangfan{30.95} & \ouyangfan{31.18} & \ouyangfan{31.21} & \ouyangfan{31.10} & \ouyangfan{30.95} \\
        \hline
        \multirow{2}{*}{\ouyangfan{$\Theta_c$}} 
        & \ouyangfan{Value} & \ouyangfan{0.3} & \ouyangfan{0.4} & \textbf{\ouyangfan{0.5}} & \ouyangfan{0.6} & \ouyangfan{0.7} & \ouyangfan{0.8} \\
        \cline{2-8}
        & \ouyangfan{Acc (\%)} & \ouyangfan{31.20} & \ouyangfan{31.21} & \ouyangfan{31.21} & \ouyangfan{31.20} & \ouyangfan{31.20} & \ouyangfan{31.19} \\
        \hline
        \multirow{2}{*}{\ouyangfan{$\lambda_{align}$}} 
        & \ouyangfan{Value} & \ouyangfan{0.1} & \textbf{\ouyangfan{0.2}} & \ouyangfan{0.3} & \ouyangfan{0.5} & \ouyangfan{0.8} & \ouyangfan{1.0} \\
        \cline{2-8}
        & \ouyangfan{Acc (\%)} & \ouyangfan{31.05} & \ouyangfan{31.21} & \ouyangfan{31.14} & \ouyangfan{31.10} & \ouyangfan{30.85} & \ouyangfan{30.45} \\
        \hline
        \multirow{2}{*}{\ouyangfan{$\lambda_{AU}$}} 
        & \ouyangfan{Value} & \ouyangfan{0.0} & \ouyangfan{0.5} & \textbf{\ouyangfan{1.0}} & \ouyangfan{1.5} & \ouyangfan{2.0} & \ouyangfan{3.0} \\
        \cline{2-8}
        & \ouyangfan{Acc (\%)} & \ouyangfan{30.25} & \ouyangfan{30.95} & \ouyangfan{31.21} & \ouyangfan{31.20} & \ouyangfan{31.05} & \ouyangfan{30.70} \\
        \hline
    \end{tabular}
\end{table}

\ouyangfan{The empirical results clearly indicate that the performance surfaces across all hyperparameters exhibit remarkable smoothness and consistency. Crucially, the default hyperparameter values originally identified on the CIFAR datasets remain either exactly optimal or near-optimal on Tiny-ImageNet-C. In particular, the model shows absolute desensitization to the fallback threshold $\Theta_c$, with the overall accuracy remaining virtually constant at 31.20\%--31.21\% across the entire search grid. These consistent convergence behaviors firmly validate that the core mechanisms of ProtoDCS possess excellent cross-dataset generalization, allowing it to function as a tuning-free framework across diverse open-set scenarios without requiring dataset-specific recalibration.}

\section{DIVERSITY-AWARE VISUAL CACHE: MECHANISM AND ANALYSIS}
\label{sec:ana}
This section elaborates on the \textit{Visual Cache} module ($\mathcal{C}_v$), which is critical for maintaining robust class representations without incurring high memory costs. We detail its specific update policy and analyze its computational efficiency.

\subsection{UPDATE POLICY AND ALGORITHM}
The maintenance of the visual cache $\mathcal{C}_v$ is governed by a \textit{diversity-quality trade-off} mechanism (Algorithm~\ref{alg:cache_update}), ensuring that the stored prototypes span the class manifold without absorbing noise.
Unlike traditional FIFO queues, we employ Aleatoric Uncertainty (AU) as a dynamic criterion for sample retention. Since AU captures inherent data noise (e.g., blur, occlusion), using it to arbitrate replacements (lines 6-10, 15-16) ensures that the cache progressively purifies the class representation, rejecting outliers even if they are semantically similar.

When a new candidate sample $x_{new}$ is admitted, the update logic proceeds as follows:
\begin{enumerate}

    \item \textbf{Redundancy Check (High Similarity):} 
    First, we compute the maximum cosine similarity $s_{\max}$ between $x_{new}$ and all existing samples in the class queue $\mathcal{Q}_k$.
    If $x_{new}$ is highly similar to an existing sample ($s_{\max} > \tau_{\text{sim}}$, default 0.9), it indicates redundancy. In this case, we only perform a replacement if $x_{new}$ has lower uncertainty (better quality) than the existing sample. This ensures that the prototype $\bm{f}_{v,k}$ (mean of the queue) progressively becomes a cleaner representation of the class center.
    


    \item \textbf{Diversity Expansion (Low Similarity):} 
    If $x_{new}$ is distinct from existing samples ($s_{\max} \le \tau_{\text{sim}}$), it contributes new semantic information.
    \begin{itemize}
        \item If the queue is not full, $x_{new}$ is directly appended.
        \item If the queue is full, we identify the "worst" sample in the current queue—defined as the one with the highest Aleatoric Uncertainty (lowest quality)—and replace it with $x_{new}$. 
    \end{itemize}
\end{enumerate}
This mechanism ensures that the cache $\mathcal{C}_v$ evolves to span the diverse visual variations of the class while actively filtering out noisy outliers.

\begin{algorithm}[h!]
    \caption{Diversity-Aware Cache Update Logic}
    \label{alg:cache_update}
    \begin{algorithmic}[1]
        \STATE \textbf{Input:} New sample feature $\bm{f}_{\text{new}}$, its initial logits $\bm{l}_{\text{new}}$, target class queue $\mathcal{Q}_k$.
        \STATE \textbf{Hyperparameters:} Queue capacity $N_{\text{cache}}$, similarity threshold $\tau_{\text{sim}}=0.9$.
        \STATE \textbf{Output:} The updated class queue $\mathcal{Q}_k$.
        
        \STATE $q_{\text{new}} \leftarrow \text{AU}(\bm{l}_{\text{new}})$. \quad \textcolor{gray}{// Calculate new sample's quality}
        
        \STATE $s_{\max} \leftarrow \max_{(\bm{f}_i, \cdot) \in \mathcal{Q}_k} \cos(\bm{f}_{\text{new}}, \bm{f}_i)$ \quad \textcolor{gray}{// Find max similarity $s_{\max}$ with items in $\mathcal{Q}_k$}
        \IF{$s_{\max} > \tau_{\text{sim}}$}
        

            \STATE $(\bm{f}_{\text{sim}}, \bm{l}_{\text{sim}}) \leftarrow \mathop{\arg\max}_{(\bm{f}_i, \bm{l}_i) \in \mathcal{Q}_k} \cos(\bm{f}_{\text{new}}, \bm{f}_i)$ \quad \textcolor{gray}{// Let $(\bm{f}_{\text{sim}}, \bm{l}_{\text{sim}})$ be the most similar item}
            
            \IF{$q_{\text{new}} < \text{AU}(\bm{l}_{\text{sim}})$}
                \STATE Replace $(\bm{f}_{\text{sim}}, \bm{l}_{\text{sim}})$ with $(\bm{f}_{\text{new}}, \bm{l}_{\text{new}})$.
            \ENDIF
        \ELSE
            \IF{$|\mathcal{Q}_k| < N_{\text{cache}}$}
                \STATE Add $(\bm{f}_{\text{new}}, \bm{l}_{\text{new}})$ to $\mathcal{Q}_k$.
            \ELSE
                \STATE Find $(\bm{f}_{\text{worst}}, \bm{l}_{\text{worst}}) \leftarrow \arg\max_{(\bm{f}_i, \bm{l}_i) \in \mathcal{Q}_k} \text{AU}(\bm{l}_i)$
                \STATE Replace $(\bm{f}_{\text{worst}}, \bm{l}_{\text{worst}})$ with $(\bm{f}_{\text{new}}, \bm{l}_{\text{new}})$ in $\mathcal{Q}_k$.
            \ENDIF
        \ENDIF
    \end{algorithmic}
\end{algorithm}

\subsection{\ouyangfan{COLD-START CACHE QUALITY AND ADAPTIVE SAMPLE REPLACEMENT}}
\begin{figure}[htbp]
    \centering
    \includegraphics[width=0.85\linewidth]{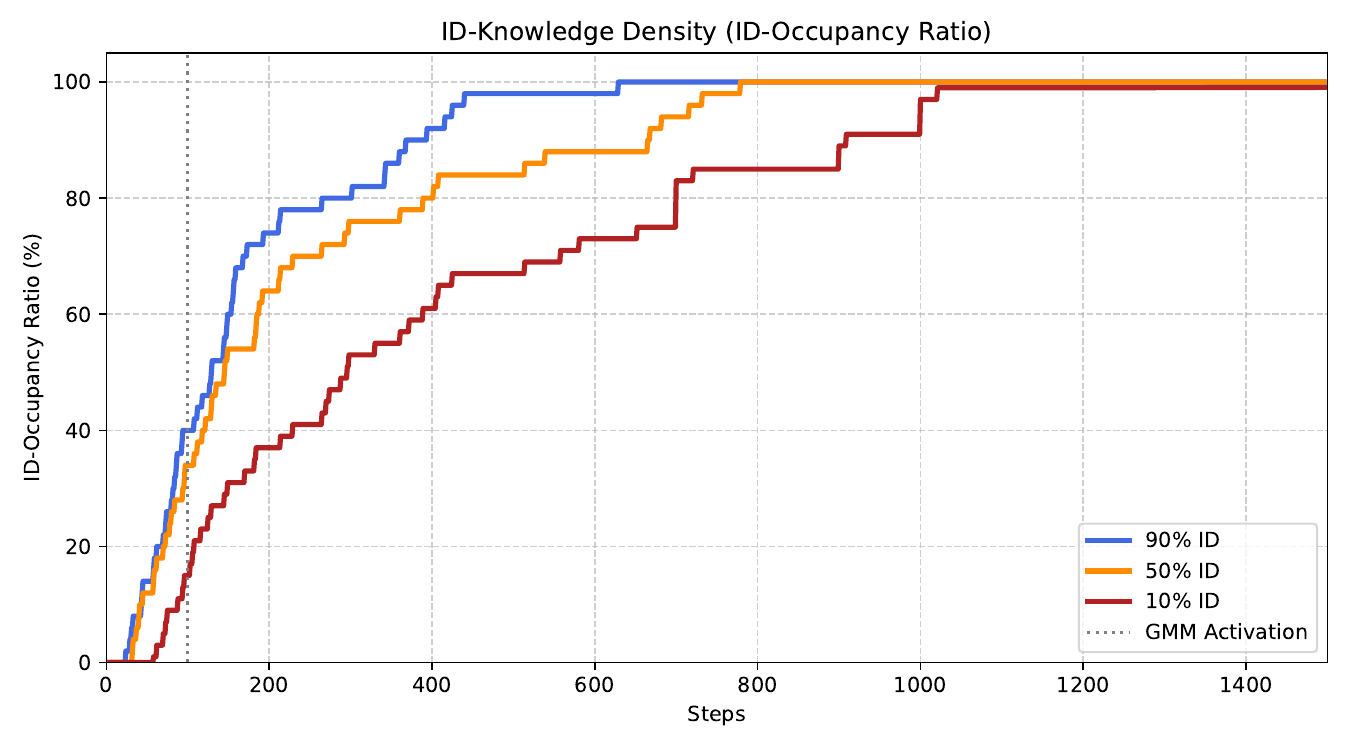}
    \caption{\ouyangfan{Dynamic evolution of the ID-Occupancy dynamics (cache quality) over 1500 test stream steps under different ID:OOD ratios (average of 15 corruptions on CIFAR-10-C). The vertical dotted line indicates the GMM activation point (step 100).}}
    \label{fig:cache_purity_sm}
\end{figure}

\ouyangfan{To evaluate the robustness of our dynamic percentile thresholds and the adaptive sample replacement capability of the visual cache during the cold-start phase ($n_{proc} < 100$) before the GMM is activated, we systematically analyze the cache quality under highly skewed data streams. We define a quantitative metric, namely the ID-Occupancy Ratio, to measure the density of genuine in-distribution (csID) knowledge stored within the cache over time:}
\begin{equation}
ID\text{-}Occupancy(t) = \frac{|x \in \mathcal{C}_{v,t} \cap x \in csID|}{Cache\ Total\ Capacity} \times 100\%,
\label{eq:id_occupancy}
\end{equation}
\ouyangfan{where $Cache\ Total\ Capacity = K \times N_{cache}$. We evaluate this metric on CIFAR-10-C under three different ID:OOD ratios: 90\%:10\% (high ID density), 50\%:50\% (medium skew), and 10\%:90\% (extreme OOD skew), with the results plotted in Fig.~\ref{fig:cache_purity_sm}.}

\ouyangfan{As illustrated in Fig.~\ref{fig:cache_purity_sm}, during the first 100 steps (the cold-start phase before GMM activation), the visual cache purity remains relatively low, particularly under the extreme 10\%:90\% skew where the ID-Occupancy ratio is below 20\%, indicating that during the early cold-start phase under severe skew, the visual cache accumulates csID samples at a slower rate due to their absolute scarcity in the incoming stream. However, as online adaptation proceeds and GMM verification is activated, our diversity-quality trade-off update mechanism (Algorithm 2) natively drives continuous cache growth and purification. Because OOD samples naturally exhibit higher aleatoric uncertainty (AU) than high-quality csID samples, they are progressively pushed to the bottom of the cache. As subsequent true csID samples arrive, they trigger the adaptive sample replacement mechanism, automatically replacing the highest-uncertainty OOD samples with high-fidelity csID features. Consequently, the ID-Occupancy Ratio under all three distributions successfully converges to 100\%, demonstrating that ProtoDCS is robustly distribution-invariant and prevents permanent cache contamination under early-stage distribution extremes.}

\subsection{COMPUTATIONAL COMPLEXITY ANALYSIS}
A key advantage of ProtoDCS is its efficiency. Here, we theoretically analyze the overhead introduced by the visual cache mechanism to substantiate the empirical results in Table IX of the main paper.

Let $N_{\text{cache}}$ be the maximum size of the queue per class (set to 5) and $K$ be the number of classes. 
The cache update involves calculating cosine similarities between the incoming sample feature $\bm{f}_{new}$ and the stored features in the target class queue $\mathcal{Q}_k$. 
Since the queue size is fixed at a small constant $N_{\text{cache}}=5$, the complexity of the similarity search and quality comparison for each incoming sample is $O(N_{\text{cache}})$, which is effectively $O(1)$. 
Crucially, this operation involves only lightweight vector dot products in the embedding dimension $d$ (e.g., 512 for ViT-B/16), avoiding any gradient backpropagation or heavy matrix multiplications associated with the backbone.
Consequently, the memory overhead is limited to storing $K \times N_{\text{cache}}$ feature vectors, and the computational latency is negligible compared to the VLM inference pass. This theoretical bound aligns with our observation that ProtoDCS maintains high throughput comparable to inference-only baselines.

\end{document}